\documentclass{article}

\usepackage{amsmath,amsthm,amssymb,amsfonts}
\usepackage{mathdef-omar}

\usepackage{enumitem}
\usepackage{cancel}
\usepackage{hyperref}
\usepackage{mathtools}
\usepackage{bbm,dsfont}

\usepackage[nonatbib,preprint]{nips_2018}

\usepackage{natbib}
\usepackage[capitalize]{cleveref}

\usepackage[textsize=tiny,
	disable
]{todonotes}
\newcommand{\todoc}[2][]{\todo[size=\scriptsize,color=blue!20!white,#1]{Csaba: #2}}
\newcommand{\todoo}[2][]{\todo[size=\scriptsize,color=red!20!white,#1]{Omar: #2}}

\newtheorem{theorem}{Theorem}

\newtheorem{lemma}[theorem]{Lemma}
\newtheorem{corollary}[theorem]{Corollary}
\newtheorem{definition}[theorem]{Definition}

\DeclareMathOperator*{\argmin}{arg\,min}

\newcommand{\dee}{\hspace{2pt} d}
\newcommand{\KL}{\operatorname{\mathit{KL}}}
\newcommand{\A}{\operatorname{\mathsf{A}}}

\newcommand{\E}{\mathbb{E}}

\newcommand{\PEW}{P@EW\xspace} 
\newcommand{\PO}{P@O\xspace} 

\usepackage{xspace}
\newcommand{\pim}{\textsc{pim}\xspace}
\newcommand{\rin}{\textsc{rin}\xspace}

\title{PAC-Bayes bounds for stable algorithms with instance-dependent priors}
\author{
Omar Rivasplata \\ UCL \And
Emilio Parrado-Hern\'{a}ndez \\ University Carlos III of Madrid \And
John Shawe-Taylor \\ UCL \And
Shiliang Sun \\  East China Normal University \And
Csaba Szepesvari  \\ Deepmind
}

\begin{document}  

\iftrue
\maketitle
\fi

\setlength{\marginparwidth}{10ex}

\begin{abstract}
PAC-Bayes bounds have been proposed to get risk estimates based on a training sample. In this paper the PAC-Bayes approach is combined with stability of the hypothesis learned by a Hilbert space valued algorithm. The PAC-Bayes setting is used with a Gaussian prior centered at the expected output. Thus a novelty of our paper is using priors defined in terms of the data-generating distribution. Our main result estimates the risk of the randomized algorithm in terms of the hypothesis stability coefficients. We also provide a new bound for the SVM classifier, which is compared to other known bounds experimentally. Ours appears to be the first stability-based bound that evaluates to non-trivial values.
\end{abstract}

\section{Introduction}
This paper combines two directions of research: stability of learning algorithms,
and PAC-Bayes bounds for algorithms that randomize with a data-dependent distribution.
The combination of these ideas enables the development of risk 
bounds that exploit stability of the learned hypothesis but are
independent of the complexity of the hypothesis class. 
We use the PAC-Bayes setting with 
`priors' defined in terms of the data-generating distribution,
as introduced by \citet{Catoni2007} 
and developed further e.g. by \citet{Lever-etal2010} and
\citet{PASTS2012}. 
Our work can be viewed as deriving specific results for this approach 
in the case of stable Hilbert space valued algorithms.

The analysis introduced by 
\citet{BE2002Stability}, 
which followed and extended \citet{LugosiPawlak1994posterior}
and was further developed by 
\citet{CelisseGuedj2016stability},
\citet{KarimCsaba2017Apriori} and 
\citet{Liu-etal2017BarcelonaPaper} among others,
shows that 
stability of learning algorithms can be used to give bounds on 
the generalisation of the learned functions. 
Their results work by assessing how small changes in the training set 
affect 
the resulting classifiers. 
Intuitively, this is because stable learning 
should ensure that slightly different training sets give similar solutions. 

In this paper we focus on the sensitivity coefficients (see our \cref{def:hyp-stab})
of the hypothesis learned by a Hilbert space valued algorithm, 
and provide an analysis leading to a PAC-Bayes bound for randomized classifiers 
under Gaussian randomization.
As a by-product of the stability analysis we derive a concentration inequality
for the output of a Hilbert space valued algorithm.
Applying it to Support Vector Machines 
\citep{ShaweCristianini2004KernelMethods,SteinwartChristmann2008SVMs}
we deduce a concentration bound for the SVM weight vector,
and also a PAC-Bayes performance bound for SVM with Gaussian randomization. 
Experimental results compare our new bound with other stability-based bounds,
and with a more standard PAC-Bayes bound.

Our work contributes to a line of research aiming to 
develop `self-bounding algorithms'
(\citet{Freund1998self-bounding}, \citet{LangfordBlum2003self-bounding})
in the sense that besides producing a predictor 
the algorithm also creates a performance certificate 
based on the available data.


\section{Main Result(s)}


We consider a learning problem where
the learner observes pairs $(X_i,Y_i)$ of patterns (inputs) $X_i$ 
from the space\footnote{All spaces where random variables
take values are assumed to be measurable spaces.}
$\mathcal{X}$ and labels $Y_i$ in the space $\mathcal{Y}$.
A training set (or sample) is a finite sequence
$S_n = ((X_1, Y_1), \ldots, (X_n, Y_n))$ of such observations.
Each pair $(X_i, Y_i)$ is a random element of $\mathcal{X}\times\mathcal{Y}$ 
whose (joint) probability law is%
\footnote{$M_1(\mathcal{Z})$ denotes the set of all probability measures 
over the space $\mathcal{Z}$.}
$P \in M_1(\mathcal{X}\times\mathcal{Y})$.
We think of $P$ as the underlying `true' (but unknown) data-generating distribution.
Examples are i.i.d. (independent and identically distributed) in the sense that
the joint distribution of $S_n$
is the $n$-fold product measure $P^{n} = P \otimes \cdots \otimes P$.

A learning algorithm is a function 
$\A : \cup_{n} (\mathcal{X}\times\mathcal{Y})^n \to \mathcal{Y}^{\mathcal{X}}$
that maps training samples (of any size) 
to predictor functions.
Given $S_n$, the algorithm produces a learned hypothesis 
$\A({S_n}): \mathcal{X} \to \mathcal{Y}$ that will be used 
to predict the label of unseen input patterns $X \in \mathcal{X}$.
%
Typically $\mathcal{X} \subset \mathbb{R}^d$ and $\mathcal{Y}\subset \mathbb{R}$.
For instance, $\mathcal{Y}=\{-1,1\}$ for binary classification, 
and $\mathcal{Y} = \mathbb{R}$ 
for regression.
A loss function $\ell : \R\times\mathcal{Y} \to [0,\infty)$
is used to assess the quality of hypotheses 
$h : \mathcal{X} \to \mathcal{Y}$. 
Say if a pair $(X,Y)$ is sampled, 
then $\ell(h(X),Y)$ quantifies the dissimilarity between
the label $h(X)$ predicted by $h$, and the actual label $Y$.
We may write $\ell_{h}(X,Y) = \ell(h(X),Y)$ to express the losses 
(of $h$)
as function of the training examples.
The (theoretical) \emph{risk} of hypothesis $h$ under data-generating distribution $P$ 
is\footnote{
Mathematicians write $\ip{f}{\nu} \stackrel{\mathrm{def}}{=} 
\int_{\mathcal{X}\times\mathcal{Y}} f(x,y) \dee \nu(x,y)$
for the integral of a function $f : \mathcal{X}\times\mathcal{Y} \to \R$ 
with respect to a (not necessarily probability) measure $\nu$ on $\mathcal{X}\times\mathcal{Y}$.
}
$R(h,P) = \ip{\ell_{h}}{P}$.
It is also called the \emph{error} 
of $h$ under $P$.
The \emph{empirical risk} of $h$ on a sample $S_n$ is 
$R(h,P_n) = \ip{\ell_{h}}{P_n}$
where $P_n = \frac{1}{n}\sum_{i=1}^{n} \delta_{(X_i,Y_i)}$ is
the empirical measure%
\footnote{
Integrals with respect to $P_n$ evaluate as follows:
$\int_{\mathcal{X}\times\mathcal{Y}} \ell(c(x),y) \dee P_n(x,y)
= \frac{1}{n} \sum_{i=1}^{n} \ell(c(X_i),Y_i)$.
}
on $\mathcal{X}\times\mathcal{Y}$ associated to the sample.
%
Notice that the risk (empirical or theoretical) is tied to the choice of a loss function.
For instance, consider binary classification
with the 0-1 loss $\ell_{01}(y',y) = \mathbf{1}[y' \neq y]$,
where $\mathbf{1}[\cdot]$ is an indicator function 
equal to $1$ when the argument is true and equal to $0$ when the argument is false.
In this case the risk is 
$R_{01}(c,P) = P[ c(X) \neq Y]$,
i.e. the probability of misclassifying the random example $(X,Y) \sim P$ 
when using $c$;
and the empirical risk is
$R_{01}(c,P_n) = \frac{1}{n} \sum_{i=1}^{n} \mathbf{1}[c(X_i) \neq Y_i]$,
i.e. the in-sample proportion of misclassified examples.

Our main theorem concerns Hilbert space valued algorithms,
in the sense that its learned hypotheses live in a Hilbert space $\mathcal{H}$.
In this case we may use the Hilbert space norm 
$\norm{w}_{\mathcal{H}} = \sqrt{\ip{w}{w}_{\mathcal{H}}}$
to measure the difference between the hypotheses 
learned from two slightly different samples.

To shorten the notation we will write $\mathcal{Z} = \mathcal{X}\times\mathcal{Y}$.
A generic element of this space is $z = (x,y)$,
the observed examples are $Z_i = (X_i,Y_i)$ and the sample of size $n$ is
$S_n = (Z_1, \ldots, Z_n)$.

\begin{definition} 
\label{def:hyp-stab}
Consider a learning algorithm 
$\A : \cup_{n} \mathcal{Z}^n \to \mathcal{H}$
where $\mathcal{H}$ is a separable Hilbert space.
We define\footnote{
For a list $\xi_1, \xi_2, \xi_3, \ldots$ and indexes $i < j$, 
we write $\xi_{i:j} = (\xi_i, \ldots, \xi_j)$,
i.e. the segment from $\xi_i$ to $\xi_j$.
}
the \emph{hypothesis sensitivity coefficients} of $\A$ as follows:
\begin{align*} 
\beta_{n} = 
\sup_{i \in [n]} \ \sup_{z_i,z_i^{\prime}} \
\norm{\A(z_{1:i-1},z_i,z_{i+1:n}) - \A(z_{1:i-1},z_i^{\prime},z_{i+1:n})}_{\mathcal{H}}\,.
\end{align*} 
\end{definition}
This is close in spirit to what is called
``uniform stability'' in the literature, 
except that 
our definition concerns stability of the learned hypothesis itself
(measured by a distance on the hypothesis space), 
while e.g. 
\citet{BE2002Stability} deal with stability of the loss functional. 
The latter could be called
``loss stability'' (in terms of ``loss sensitivity coefficients'')
for the sake of informative names.

Writing $z_{1:n} \approx z^{\prime}_{1:n}$ when
these $n$-tuples differ at one entry (at most), 
an equivalent formulation to the above is 
$\beta_{n} = 
\sup_{z_{1:n} \approx z^{\prime}_{1:n}} \
\norm{\A(z_{1:n}) - \A(z^{\prime}_{1:n})}_{\mathcal{H}}$.
In particular, if two samples
$S_n$ and $S^{\prime}_{n}$ 
differ only on one example,
then $\norm{\A(S_{n})-\A(S^{\prime}_{n})}_{\mathcal{H}} \leq \beta_n$.
Thus our definition implies stability with respect to 
replacing one example with an independent copy. 
Alternatively, one could define
$\beta_{n} = 
\esssup_{S_{n} \approx S^{\prime}_{n}} \
\norm{\A(S_{n}) - \A(S^{\prime}_{n})}_{\mathcal{H}}$,
which corresponds to the ``uniform argument stability'' of \citet{Liu-etal2017BarcelonaPaper}.
We avoid the `almost-sure' technicalities by defining our $\beta_n$'s 
as the maximal difference (in norm) with respect to all $n$-tuples 
$z_{1:n} \approx z^{\prime}_{1:n}$.
The extension to sensitivity when changing several examples is natural: 
$\norm{\A(z_{1:n}) - \A(z^{\prime}_{1:n})}_{\mathcal{H}}
\leq \beta_n \sum_{i=1}^{n}\mathbf{1}[z_{i} \neq z^{\prime}_{i}]$.
Note that $\beta_n$ is a Lipschitz factor with respect to the
Hamming distance. 
The ``total Lipschitz stability'' of \citet{Kontorovich2014concentration} 
is a similar notion for stability of the loss functional.
The ``collective stability'' of \citet{London-etal2013}
is not comparable to ours (different setting) despite the similar look.

We will consider randomized classifiers that operate as follows.
Let $\mathcal{C}$ be the classifier space, and
let $Q \in M_1(\mathcal{C})$ be a probability distribution over the classifiers.
To make a prediction the randomized classifier picks $c \in \mathcal{C}$
according to $Q$ and predicts a label with the chosen $c$.
Each prediction is made with a fresh $c$ draw.
For simplicity
we use the same label $Q$ for the probability distribution 
and for the corresponding randomized classifier.
The risk measures $R(c,P)$ and $R(c,P_n)$ are extended
to randomized classifiers:
$R(Q,P) \equiv \int_{\mathcal{C}} R(c,P) \dee Q(c)$ is the
\emph{average theoretical risk} of $Q$,
and $R(Q,P_n) \equiv \int_{\mathcal{C}} R(c,P_n) \dee Q(c)$
its \emph{average empirical risk}.
Given two distributions $Q_{0}, Q \in M_1(\mathcal{C})$,
the Kullback-Leibler divergence (a.k.a. relative entropy) 
of $Q$ with respect to $Q_{0}$ is
\begin{align*}
\vspace{-4mm}
\KL(Q \Vert Q_{0}) = \int_{\mathcal{C}} \log\bigl( \frac{dQ}{dQ_{0}} \bigr) \dee Q\,.
\vspace{-4mm}
\end{align*} 
Of course this 
makes sense when $Q$ is absolutely continuous with respect to $Q_{0}$,
which ensures that the Radon-Nikodym derivative $dQ/dQ_{0}$ exists.
For Bernoulli distributions with parameters $q$~and~$q_0$ we write
$\KL(q \Vert q_0) = q \log(\frac{q}{q_0}) + (1-q)\log(\frac{1-q}{1-q_0})$,
and
$\KL_{+}(q \Vert q_0) = \KL(q \Vert q_0) \mathbf{1}[q<q_0]$.

\subsection{A PAC-Bayes bound for stable algorithms with Gaussian randomization}
This is our main result:
\begin{theorem}
\label{thm:main}
Let $\A$ be a Hilbert space valued algorithm.
Suppose that (once trained) the algorithm will randomize according to
Gaussian distributions $Q = \mathcal{N}(\A(S_n), \sigma^2 I)$.
If $\A$ has hypothesis stability coefficients $\beta_n$,
then for any randomization variance $\sigma^2 > 0$,
for any $\delta \in (0,1)$, with probability $\geq 1-2\delta$ we have %
\[
\KL_{+}( R(Q,P_n) \Vert R(Q,P) ) \leq
\frac{ \frac{n\beta_n^2}{2\sigma^2} 
			\left( 1 + \sqrt{\frac{1}{2}\log\bigl(\frac{1}{\delta}\bigr)} \ \right)^2
			+ \log(\frac{n+1}{\delta})
     }{n} \,.
\]
\end{theorem} 

The proof of our theorem combines stability of the learned hypothesis
(in the sense of \cref{def:hyp-stab}) and a PAC-Bayes bound
for the average theoretical error of a randomized classifier,
quoted below in \cref{s:proofs} (Proofs) for reference.
Note that the randomizing distribution $Q$ depends on the sample.
Literature \todoo{work on literature}
on the PAC-Bayes framework for learning linear classifiers include 
\citet{Germain-etal2015} and \citet{PASTS2012} with references.
Application of the PAC-Bayes framework to training neural networks
can be seen e.g. in \citet{London2017}, \cite{DziugaiteRoy2017}.

\subsection{A PAC-Bayes bound for SVM with Gaussian randomization}

For a Support Vector Machine (SVM) with feature map $\varphi : \mathcal{X} \to \mathcal{H}$
into a separable Hilbert space $\mathcal{H}$,
we may 
identify\footnote{Riesz representation theorem is behind this identification.}
a linear classifier $c_{w}(\cdot) = \sign(\langle w, \varphi(\cdot) \rangle)$
with a vector $w \in \mathcal{H}$. %
With this identification we can regard an SVM as a 
Hilbert space\footnote{$\mathcal{H}$ may be infinite-dimensional (e.g. Gaussian kernel).}
valued mapping
that based on a training sample $S_n$ learns a weight vector 
$W_n = \operatorname{SVM}(S_n) \in \mathcal{H}$.
In this context, stability of the SVM's solution 
then reduces to stability of the learned weight vector.

To be specific, let $\operatorname{SVM}_{\lambda}(S_n)$ be the SVM
that regularizes the empirical risk over the sample $S_n$
by solving the following optimization problem:
\begin{equation*}
\argmin_{w} \biggl( \frac{\lambda}{2} \norm{w}^2 
	+ \frac{1}{n}\sum_{i=1}^{n} \ell(c_{w}(X_i),Y_i) \biggr)\,.
\tag{svm}
\label{eq:svm}
\end{equation*}

Our stability coefficients in this case satisfy 
$\beta_n \leq \frac{2}{\lambda n}$
(Example 2 of \citet{BE2002Stability}, adapted to our setting).
Then a direct application of our \cref{thm:main} 
together with a concentration argument for the SVM weight vector
(see our \cref{concentration-SVM} below)
gives the following:

\begin{corollary}
\label{pbb-SVM}
Let $W_n = \operatorname{SVM}_{\lambda}(S_n)$.
Suppose that (once trained) the algorithm will randomize according to
Gaussian\footnote{See \cref{app:gaussian} about the interpretation of
Gaussian randomization for a Hilbert space valued algorithm.}
distributions $Q = \mathcal{N}(W_n, \sigma^2 I)$.
For any randomization variance $\sigma^2 > 0$, for any $\delta \in (0,1)$,
with probability $\geq 1-2\delta$ we have
\[
\KL_{+}( R(Q,P_n) \Vert R(Q,P) ) 
\leq
\frac{ \frac{2}{\sigma^2\lambda^2 n} 
			\left( 1 + \sqrt{\frac{1}{2}\log\bigl(\frac{1}{\delta}\bigr)} \ \right)^2
			+ \log(\frac{n+1}{\delta})
     }{n} \,.
\]
\end{corollary}

In closing this section we mention that our main theorem is general
in that it may be specialized to any Hilbert space valued algorithm.
This covers any regularized ERM algorithm
\citep{Liu-etal2017BarcelonaPaper}.
We applied it to SVM's whose hypothesis sensitivity coefficients 
(as in our \cref{def:hyp-stab}) are known.
It can be argued that neural networks (NN's) 
fall under this framework as well.
Then an appealing future research direction, with deep learning in view,
is to figure out the sensitivity coefficients of NN's trained by 
Stochastic Gradient Descent.
Then our main theorem could be applied to provide non-vacuous bounds 
for the performance of NN's, which we believe is very much needed.

\section{Comparison to other bounds}
\label{s:bounds-to-compare}

For reference we list several risk bounds (including ours).
They are in the context of binary classification ($\mathcal{Y} = \{ -1,+1 \}$).
For clarity, risks under the 0-1 loss are denoted by $R_{01}$ 
and risks with respect to the (clipped) hinge loss are denoted by $R_{\mathrm{hi}}$.
Bounds requiring a Lipschitz loss function do not apply to the 0-1 loss.
However, the 0-1 loss is upper bounded by the hinge loss, allowing us
to upper bound the risk with respect to the former in therms of the risk
with respect to the latter. 
On the other hand, results requiring a bounded loss function do not apply
to the regular hinge loss. In those cases the clipped hinge loss is used,
which enjoys boundedness and Lipschitz continuity.

\subsection{\PEW: Our new instance-dependent PAC-Bayes bound}

Our \cref{pbb-SVM}, 
with $Q = \mathcal{N}(W_n, \sigma^2 I)$,
a Gaussian centered at $W_n = \operatorname{SVM}_{\lambda}(S_n)$ 
with randomization variance $\sigma^2$,
gives the following risk bound
which holds with probability $\geq 1-2\delta$:
\begin{align*}
\KL_{+}&( R_{01}(Q,P_n) \Vert R_{01}(Q,P) ) 
\leq \frac{2}{\sigma^2 \lambda^2 n^2} 
	\biggl( 1 + \sqrt{\frac{1}{2}\log\bigl(\frac{1}{\delta}\bigr)} \ \biggr)^2
		+ \frac{1}{n}\log\bigl(\frac{n+1}{\delta}\bigr) \,.
\end{align*}

As will be clear from the proof (see \cref{s:proofs} below), this bound is obtained
from the PAC-Bayes bound (\cref{thm:pbb}) using a prior centered at the expected weight.

\subsection{\PO: Prior at the origin PAC-Bayes bound}

The PAC-Bayes bound \cref{thm:pbb} 
again with $Q = \mathcal{N}(W_n, \sigma^2 I)$,
gives the following risk bound
which holds with probability $\geq 1-\delta$:
\begin{align*}
\forall \sigma^2,\hspace{3mm} 
\KL_{+}( R_{01}(Q,P_n) \Vert R_{01}(Q,P) ) 
\leq  \frac{1}{2 \sigma^2 n} \norm{W_n}^2
	+ \frac{1}{n} \log\bigl(\frac{n+1}{\delta}\bigr) \,.
\end{align*}

\subsection{Bound of \citet{Liu-etal2017BarcelonaPaper}}

From Corollary 1 of \citet{Liu-etal2017BarcelonaPaper} 
(but with $\lambda$ as in formulation \eqref{eq:svm})
we get the following risk bound
which holds with probability $\geq 1-2\delta$:
\begin{align*}
R_{01}(W_n,P) \leq R_{\mathrm{hi}}(W_n,P) 
\leq R_{\mathrm{hi}}(W_n,P_n) 
+ \frac{\mathbf{8}}{\lambda n}\sqrt{2\log\bigl(\frac{2}{\delta}\bigr)} 
	+ \sqrt{\frac{1}{2n}\log\bigl(\frac{1}{\delta}\bigr)} \,.
\end{align*}


We use Corollary 1 of \citet{Liu-etal2017BarcelonaPaper} 
with $B=1$, $L=1$ and $M = 1$ (clipped hinge loss).

\subsection{Bound of \citet{BE2002Stability}}

From Example 2 of \citet{BE2002Stability} 
(but with $\lambda$ as in formulation \eqref{eq:svm})
we get the following risk bound
which holds with probability $\geq 1-\delta$:
\begin{align*}
R_{01}(W_n,P) \leq R_{\mathrm{hi}}(W_n,P) 
\leq R_{\mathrm{hi}}(W_n,P_n) 
+ \frac{\mathbf{2}}{\lambda n}
+ \Bigl(1+\frac{\mathbf{4}}{\lambda}\Bigr) \sqrt{\frac{1}{2n}\log\bigl(\frac{1}{\delta}\bigr)} \,.
\end{align*}


We use Example 2 and Theorem 17 (based on Theorem 12) of \citet{BE2002Stability}
with $\kappa=1$ (normalized kernel) and $M=1$ (clipped hinge loss).

In \cref{app:SVMformulations} below there is a list of different SVM formulations,
and how to convert between them. We found it useful when implementing code for experiments.


There are obvious differences in the nature of these bounds:
the last two (\citet{Liu-etal2017BarcelonaPaper} and \citet{BE2002Stability})
are risk bounds for the (un-randomized) classifiers,
while the first two (\PEW, \PO) give an upper bound
on the KL-divergence between the average risks (empirical to theoretical)
of the randomized classifiers. Of course inverting the KL-divergence
we get a bound for the average theoretical risk in terms of the
average empirical risk and the (square root of the) right hand side.
Also, the first two bounds have an extra parameter,
the randomization variance ($\sigma^2$), that can be optimized.
Note that \PO bound is not based on stability, 
while the other three bounds are based on stability notions. 
Next let us comment on how these bounds compare quantitatively.

Our \PEW bound and the \PO bound are similar except for
the first term on the right hand side. This term comes from the KL-divergence
between the Gaussian distributions. Our \PEW bound's first term improves with 
larger values of $\lambda$, which in turn penalize the norm of the weight vector 
of the corresponding SVM, resulting in a small first term in \PO bound.
Note that \PO bound is equivalent
to the setting of $Q = \mathcal{N}(\mu W_n/\norm{W_n}, \sigma^2 I)$, a Gaussian
with center in the direction of $W_n$, at distance $\mu > 0$ from the origin
(as discussed in \citet{Langford05} and implemented in \citet{PASTS2012}).

The first term on the right hand side of our \PEW bound
comes from the concentration of the weight (see our \cref{concentration-SVM}).
Lemma 1 of \citet{Liu-etal2017BarcelonaPaper}
implies a similar concentration inequality for the weight vector,
but it is not hard to see that our concentration bound is slightly better.

Finally, in the experiments we compare our \PEW bound with \citet{BE2002Stability}.

\section{Proofs}
\label{s:proofs}

As we said before, the proof of our \cref{thm:main}
combines stability of the learned hypothesis
(in the sense of our \cref{def:hyp-stab}) and a well-known PAC-Bayes bound,
quoted next for reference:


\begin{theorem}{\em (PAC-Bayes bound)\/}
\label{thm:pbb}
Consider a learning algorithm 
$\A : \cup_{n} (\mathcal{X}\times\mathcal{Y})^n \to \mathcal{C}$.
For any $Q_{0} \in M_1(\mathcal{C})$, 
and for any $\delta \in (0,1)$,
with probability $\geq 1-\delta$ we have
\[
\forall Q \in M_1(\mathcal{C}),\hspace{2mm} 
\KL_{+}( R(Q,P_n) \Vert R(Q,P) ) \leq
\frac{\KL(Q \Vert Q_{0})+\log(\frac{n+1}{\delta})}{n} \,.
\]
The probability is over the generation of the training sample $S_{n} \sim P^{n}$.
\end{theorem} 
The above is Theorem 5.1 of \citet{Langford05},
though see also Theorem 2.1 of \citet{Germain-etal2009}.
To use the PAC-Bayes bound, 
we will use $Q_0 = \mathcal{N}(\E[\A(S_n)], \sigma^2 I)$ 
and $Q = \mathcal{N}(\A(S_n), \sigma^2 I)$,
a Gaussian distribution centered at the expected output 
and a Gaussian (posterior) distribution centered at the random output $\A(S_n)$,
both with covariance operator $\sigma^2 I$.
The KL-divergence between those Gaussians scales with
$\norm{\A(S_n) - \E[\A(S_n)]}^2$. More precisely:
\begin{align*}
\KL( Q \Vert Q_{0} )
= \frac{1}{2 \sigma^2} \norm{\A(S_n) - \E[\A(S_n)]}^2 \,.
\end{align*}
Therefore, bounding $\norm{\A(S_n) - \E[\A(S_n)]}$ will give 
(via the PAC-Bayes bound of \cref{thm:pbb} above)
a corresponding bound on 
the divergence between the average empirical risk $R(Q,P_n)$
and the average theoretical risk $R(Q,P)$ of the randomized classifier $Q$.
Hypothesis stability (in the form of our \cref{def:hyp-stab}) 
implies a concentration inequality for $\norm{\A(S_n) - \E[\A(S_n)]}$.
This is done in our \cref{concentration-StableAlg} 
(see \cref{ss:stability-to-concentration} below)
and completes the circle of ideas to prove our main theorem.
The proof of our concentration inequality 
is based on an extension  of the bounded differences theorem of McDiarmid 
to vector-valued functions discussed next.

\subsection{Real-valued functions of the sample}
\label{ss:bd-real} 

To shorten the notation let's present the training sample as
$S_n = (Z_1,\ldots,Z_n)$ where each example $Z_i$ is a random variable
taking values in the (measurable) space $\mathcal{Z}$.
We quote a well-known theorem:

\begin{theorem}{\em (McDiarmid inequality) \/}
\label{thm:McDiarmid}
Let $Z_1,\ldots,Z_n$ be independent $\mathcal{Z}$-valued random variables,
and $f : \mathcal{Z}^n \to \R$ a real-valued function such that 
for each $i$ and for each list of `complementary' arguments 
$z_1,\ldots,z_{i-1},z_{i+1},\ldots,z_n$
we have
\[
\sup_{z_i,z_i^{\prime}} 
|f(z_{1:i-1},z_i,z_{i+1:n})
  -f(z_{1:i-1},z_i^{\prime},z_{i+1:n})|
\leq c_i\,.
\]
Then for every $\epsilon > 0$,
$
\operatorname{Pr}
\left\{ f(Z_{1:n}) - \E[f(Z_{1:n})] > \epsilon \right\}
\leq \exp\left( \frac{-2\epsilon^2}{\sum_{i=1}^{n} c_i^2} \right)
$.
\end{theorem}




McDiarmid's inequality applies to a \emph{real-valued} function
of independent random variables. 
Next we present an extension to \emph{vector-valued} functions
of independent random variables. The proof follows the steps of the proof
of the classic result above, but we have not found this result in the literature, hence we include the details.

\subsection{Vector-valued functions of the sample}
\label{ss:bd-vector} 

Let $Z_1,\ldots,Z_n$ be independent $\mathcal{Z}$-valued random variables
and $f : \mathcal{Z}^n \to \mathcal{H}$ 
a function into a separable Hilbert space.
We will prove that bounded differences \emph{in norm}\footnote{%
The Hilbert space norm, induced by the inner product of $\mathcal{H}$.}
implies concentration of $f(Z_{1:n})$ around its mean \emph{in norm},
i.e., that $\norm{f(Z_{1:n}) - \E f(Z_{1:n})}$ is small
with high probability.

Notice that McDiarmid's theorem can't be applied directly to 
$f(Z_{1:n}) - \E f(Z_{1:n})$ when $f$ is vector-valued.
We will apply McDiarmid to the real-valued $\norm{f(Z_{1:n}) - \E f(Z_{1:n})}$,
which will give an upper bound for $\norm{f - \E f}$ in terms of $\E\norm{f - \E f}$.
The next lemma upper bounds $\E\norm{f - \E f}$ 
for a function $f$ with bounded differences in norm.
Its proof is in \cref{app:proof-lemma}.

\begin{lemma}
\label{lemma-fbd} 
Let $Z_1,\ldots,Z_n$ be independent $\mathcal{Z}$-valued random variables,
and $f : \mathcal{Z}^n \to \mathcal{H}$ a function into a Hilbert space $\mathcal{H}$
satisfying the bounded differences property:
for each $i$ and for each list of `complementary' arguments 
$z_1,\ldots,z_{i-1},z_{i+1},\ldots,z_n$
we have
\[
\sup_{z_i,z_i^{\prime}} 
\| f(z_{1:i-1},z_i,z_{i+1:n})
  -f(z_{1:i-1},z_i^{\prime},z_{i+1:n}) \|
\leq c_i\,.
\]
Then $\E\norm{f(Z_{1:n})-\E[f(Z_{1:n})]} \leq \sqrt{\sum_{i=1}^{n} c_i^2}$.
\end{lemma}

If the vector-valued function $f(z_{1:n})$
has bounded differences in norm (as in the Lemma) and $C \in \R$ is any constant, 
then the real-valued function
$\| f(z_{1:n}) - C \|$ has the bounded differences property (as in McDiarmid's theorem).
In particular this is true for $\| f(z_{1:n}) - \E f(Z_{1:n}) \|$
(notice that $\E f(Z_{1:n})$ is constant over replacing $Z_i$ by
an independent copy $Z_i^{\prime}$)
so applying McDiarmid's inequality to it, combining with \cref{lemma-fbd},
we get the following theorem:

\begin{theorem}
\label{thm:concentration-fbd} 
Under the assumptions of \cref{lemma-fbd},
for any $\delta \in (0,1)$,
with probability $\geq 1 - \delta$ we have
\begin{align*}
\| f(Z_{1:n}) - \E[f(Z_{1:n})] \|
\leq \sqrt{\sum_{i=1}^{n} c_i^2} 
		+ \sqrt{\frac{\sum_{i=1}^{n} c_i^2}{2}\log\bigl(\frac{1}{\delta}\bigr)}\,.
\end{align*}
\end{theorem}

Notice that the vector $c_{1:n} = (c_1,\ldots,c_n)$ of difference bounds 
appears in the above inequality only through its Euclidean norm
$\norm{c_{1:n}}_2 = \sqrt{\sum_{i=1}^{n} c_i^2}$.

\subsection{Stability implies concentration}
\label{ss:stability-to-concentration}

The hypothesis sensitivity coefficients give concentration of the learned hypothesis:

\begin{corollary} 
\label{concentration-StableAlg}
Let $\A$ be a Hilbert space valued algorithm.
Suppose $\A$ has hypothesis sensitivity coefficients $\beta_{n}$.
Then for any $\delta \in (0,1)$,
with probability $\geq 1-\delta$ we have
\begin{align*}
\norm{\A(S_n) - \E[\A(S_n)]} 
\leq \sqrt{n} \ \beta_{n}
\left( 1 + \sqrt{\frac{1}{2}\log\bigl(\frac{1}{\delta}\bigr)} \right)\,.
\end{align*}
\end{corollary}

This is a consequence of \cref{thm:concentration-fbd} since
$c_{i} \leq \beta_{n}$ for $i = 1, \ldots, n$, 
hence $\norm{c_{1:n}} \leq \sqrt{n} \ \beta_{n}$.

Last (not least) we deduce concentration of the weight vector
$W_n = \operatorname{SVM}_{\lambda}(S_n)$.

\begin{corollary} 
\label{concentration-SVM}
Let $W_n = \operatorname{SVM}_{\lambda}(S_n)$.
Suppose that the kernel used by SVM is bounded by $B$.
\todoc{This was messed up. I think $B$ is the upper bound on the features.}
For any $\lambda > 0$,
for any $\delta \in (0,1)$,
with probability $\geq 1-\delta$ we have
\begin{align*}
\norm{W_n - \E[W_n]} 
\leq \frac{2B}{\lambda \sqrt{n}}
\left( 1 + \sqrt{\frac{1}{2}\log\bigl(\frac{1}{\delta}\bigr)} \right)\,.
\end{align*}
\end{corollary}

Under these conditions 
we have hypothesis sensitivity coefficients 
$\beta_{n} \leq \frac{2B}{\lambda n}$
(we follow \citet{BE2002Stability}, Example~2 and Lemma~16, 
adapted to our setting).
Then apply \cref{concentration-StableAlg}.


\section{Experiments} 
\newcommand{\sigmarbf}{\sigma} 
The purpose of the experiments was to explore the strengths and potential weaknesses of our new bound 
in relation to the alternatives presented earlier, as well as to explore the bound's ability to help model selection.
For this, to facilitate comparisons, taking the setup of  \cite{PASTS2012}, 
we experimented with the five UCI datasets described there.
However, we present results for \pim and \rin only, as the results on the other datasets mostly followed the results on these and in a way these two datasets are the most extreme. 
In particular, they are the smallest and largest with dimensions $768\times 8$ ($768$ examples, and $8$ dimensional feature space), and $7200\times 20$, respectively.

\begin{figure}[t]
\includegraphics[width=0.5\textwidth]{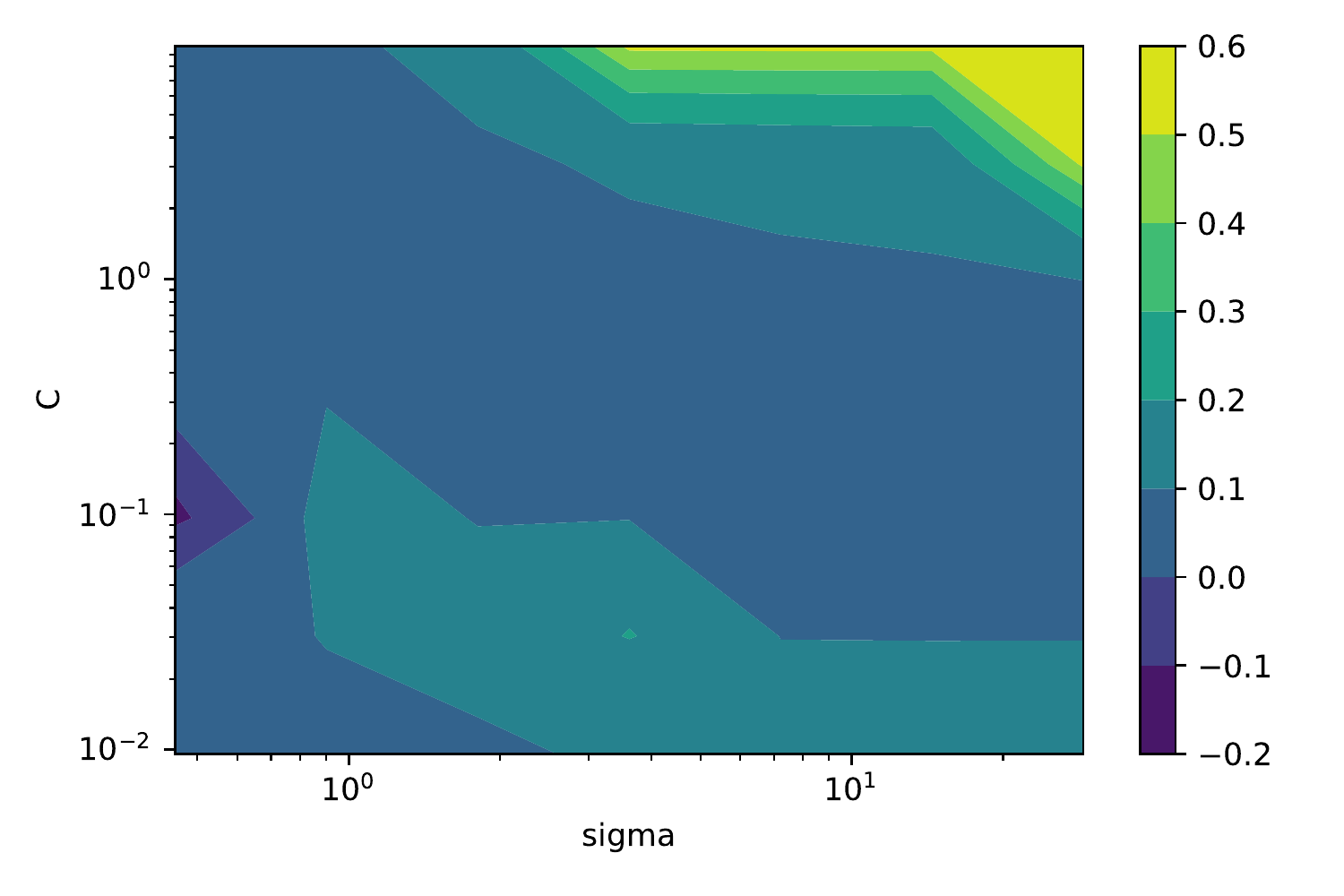}
\includegraphics[width=0.5\textwidth]{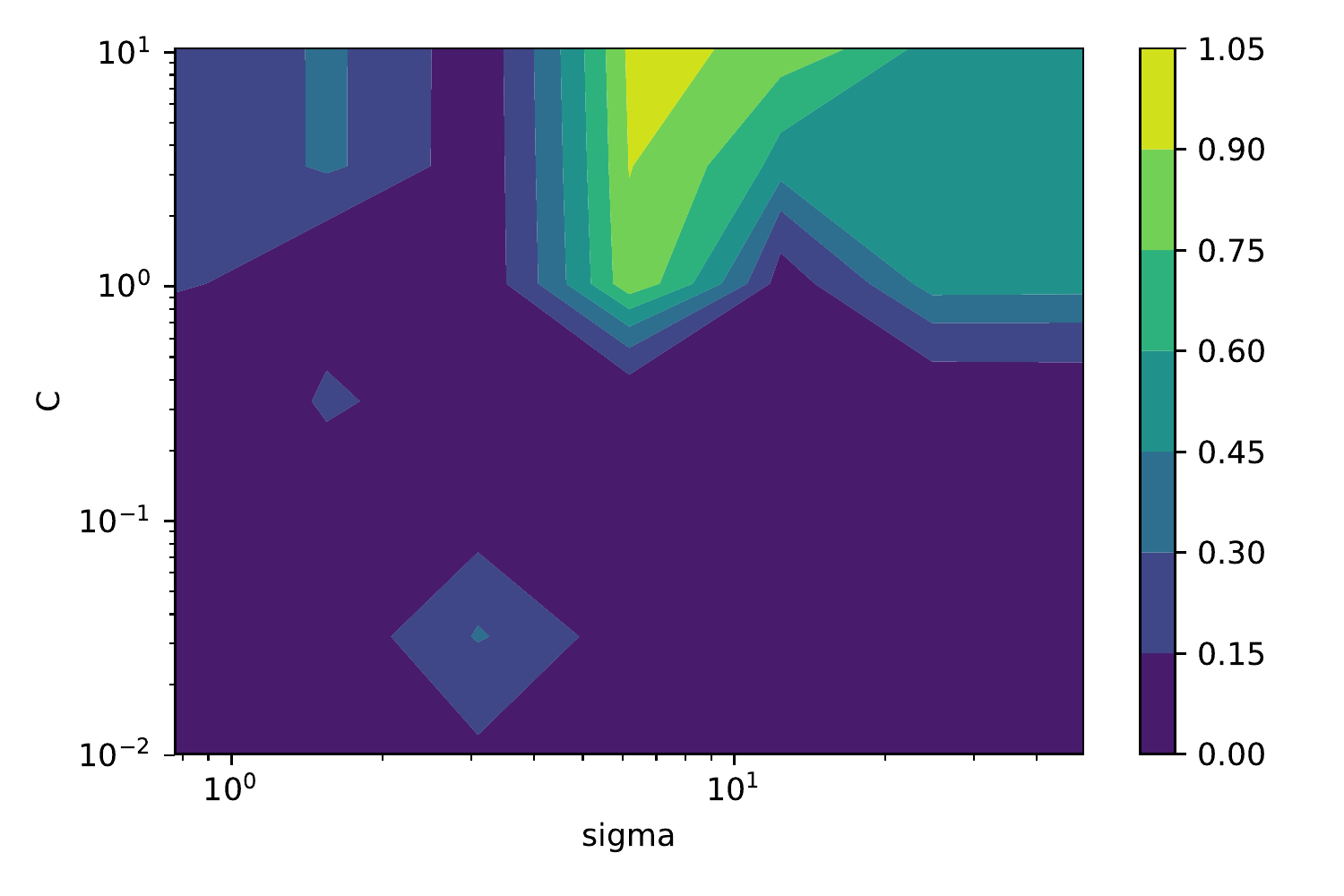}
\caption{Tightness of \PO bound on \pim (left) and \rin (right) shown as the difference between the bound and the test error of the underlying randomized classifier. Smaller values are preferred.
}
\label{fig:pootight}
\end{figure}

\begin{figure}[t]
\includegraphics[width=0.5\textwidth]{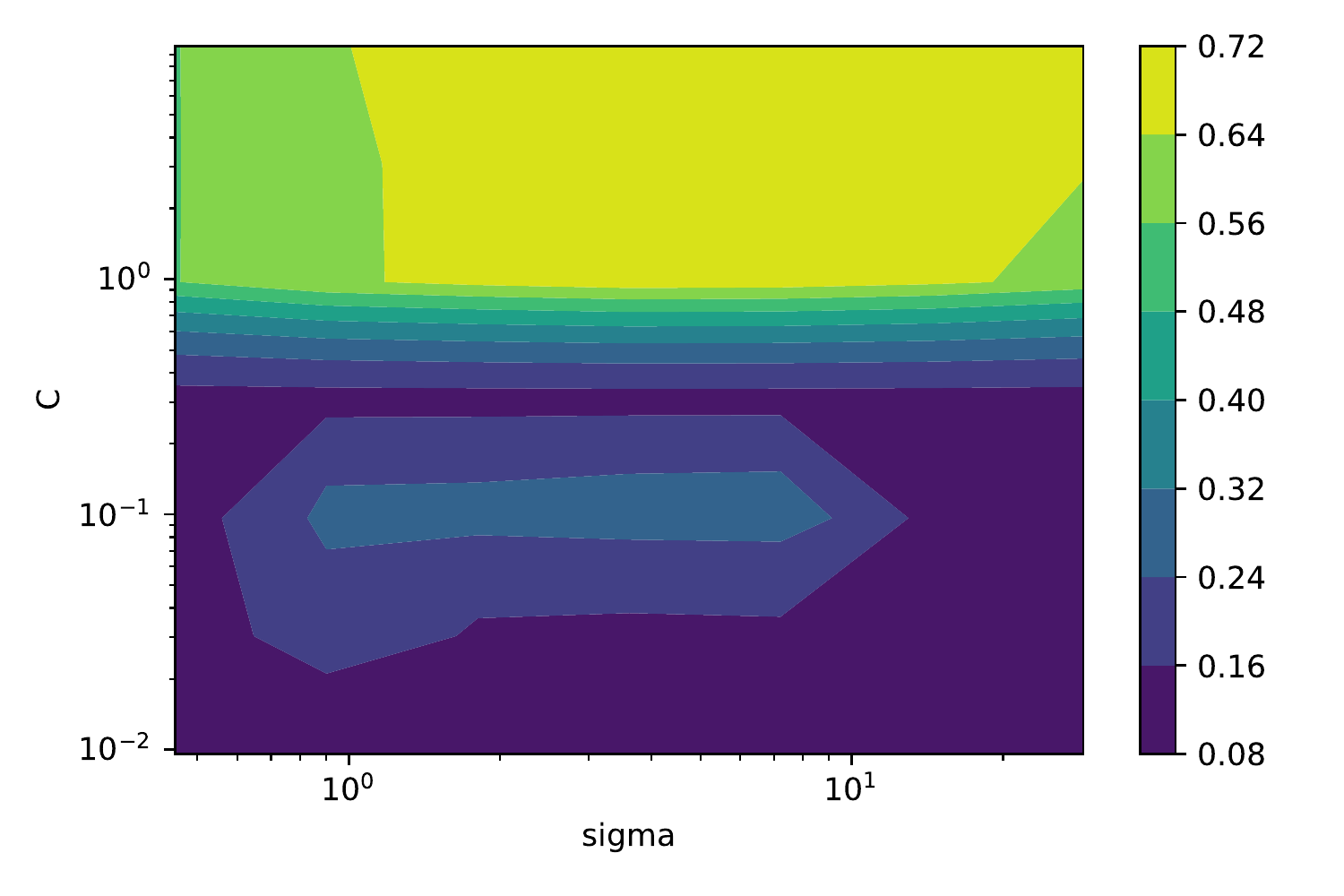}
\includegraphics[width=0.5\textwidth]{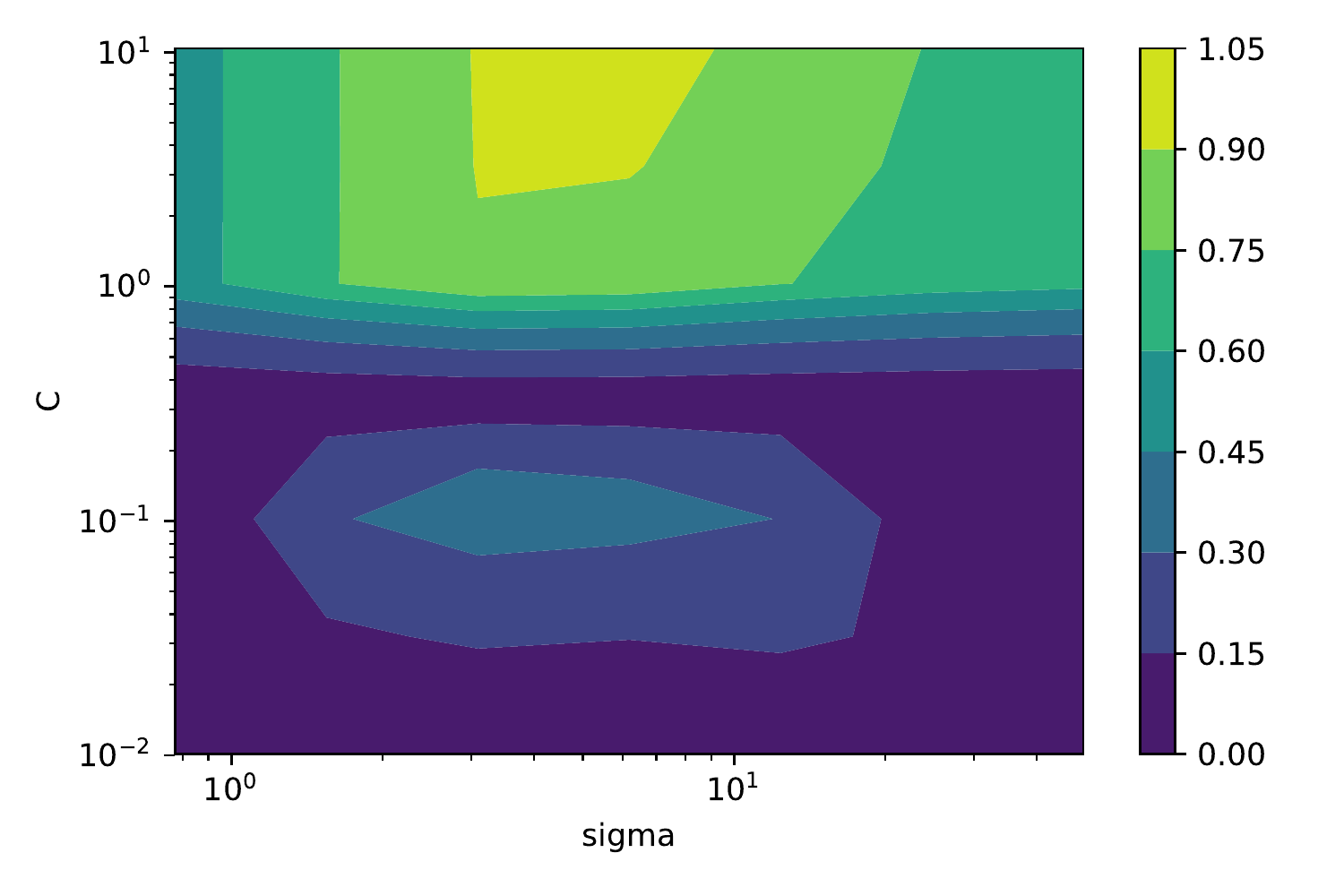}
\caption{Tightness of \PEW bound (the bound derived here) on \pim (left) and \rin (right) shown as the difference between the bound and the test error of the underlying randomized classifier. Smaller values are preferred.
}
\label{fig:oourtight}
\end{figure}

\textbf{Model and data preparation} We used an offset-free SVM classifier 
with a Gaussian RBF kernel $\kappa(x,y) = \exp(-\norm{x-y}^2_2/(2\sigmarbf^2))$ 
with RBF width parameter $\sigmarbf>0$.
The SVM used the so-called standard SVM-C formulation; 
the conversion between our and the SVM-C formulation which multiplies 
the total (hinge) loss by $C>0$ is given by $C = \frac{1}{\lambda n}$ 
where $n$ is the number of training examples 
and $\lambda>0$ is the penalty in our formulation \eqref{eq:svm}. 
The datasets were split into a training and a test set 
using the {\tt train\_test\_split } 
method of {\tt scikit}, \todoc{Ref}
keeping $80\%$ of the data for training and $20\%$ for testing. 

\textbf{Model parameters} 
Following the procedure suggested in Section~2.3.1 of \citet{ChZi05}, 
we set up a geometric $7\times 7$ grid over the $(C,\sigmarbf)$-parameter space
where $C$ ranges between $2^{-8}C_0$ and $2^2 C_0$ and $\sigmarbf$ ranges 
between $2^{-3}\sigmarbf_0$ and $2^3 \sigmarbf_0$, where 
$\sigmarbf_0$ is the median of the Euclidean distance between 
pairs of data points of the training set, and given $\sigmarbf_0$, $C_0$ is obtained 
as the reciprocal value of the empirical variance of data in feature space 
underlying the RBF kernel with width $\sigmarbf_0$.
The grid size was selected for economy of computation. 
The grid lower and upper bounds for $\sigmarbf$ were ad-hoc, though they were 
inspired by the literature, while for the same for $C$, 
we enlarged the lower range to focus on the region of the parameter space where 
the stability-based bounds have a better chance to be effective: 
In particular, the stability-based bounds grow with $C$ in a linear fashion, 
with a coefficient that was empirically observed to be close to one.

\textbf{Computations}
For each of the $(C,\sigmarbf)$ pairs on the said grid, we trained an SVM-model using a Python implementation of the SMO algorithm of \cite{Platt99}, adjusted to SVMs with no offset (\citet{Steinwart2008SVMs} argue that ``the offset term has neither a known theoretical nor an empirical advantage'' for the Gaussian RBF kernel). 
We then calculated various bounds using the obtained model, as well as the corresponding test error rates (recall that the randomized classifiers' test error is different than the test error of the SVM model that uses no randomization).
The bounds compared were the two mentioned hinge-loss based bounds:  The bound by \citet{Liu-etal2017SpanishPaper} and that of
\citet{BE2002Stability}. In addition we calculated the \PO and (our) \PEW bound. When these latter were calculated we optimized the randomization variance parameter $\sigma_{\text{noise}}^2$ by minimizing error estimate obtained from the respective bound (the KL divergence was inverted numerically). Further details of this can be found in \cref{sec:expdetails}.

\textbf{Results}
As explained earlier our primary interest is to explore the various bounds strengths and weaknesses.
In particular, we are interested in their tightness, as well as their ability to support model selection.
As the qualitative results were insensitive to the split, results for a single ``random'' (arbitrary) split are shown only.

{\em Tightness} The hinge loss based bounds gave trivial bounds over almost all pairs of $(C,\sigmarbf)$. Upon investigating this we found that this is because the hinge loss takes much larger values than the training error rate unless $C$ takes large values (cf. \cref{fig:hinge} in \cref{sec:expdetails}). However, for large values of $C$, both of the bounds are vacuous. In general, the stability based bounds (\citet{Liu-etal2017SpanishPaper,BE2002Stability} and our bound) are sensitive to large values of $C$.
\cref{fig:pootight} show the difference between the \PO bound and the test error of the underlying respective randomized classifiers
as a function of $(C,\sigmarbf)$
while \cref{fig:oourtight} shows the difference between the \PEW bound and the test error of the underlying randomized classifier.
(\cref{fig:testerrpoo,fig:testerroour} in the appendix show the test errors for these classifiers, while \cref{fig:poo,fig:oour} shows the bound.)
The meticulous reader may worry about that it appears that on the smaller dataset, \pim, the difference shown for \PO is sometimes negative. 
As it turns out this is due to the randomness of the test error: Once we add a confidence correction that accounts for the randomness of the test 
small test set ($n_{\text{test}}=154$) this difference disappears once we correct the test error for this.
From the figures the most obvious difference between the bounds is that the \PEW bound is sensitive to the value of $C$ and it becomes loose for larger values of $C$. This is expected: As noted earlier, stability based bounds, which \PEW is an instance of, are sensitive to $C$. The \PO bound shows a weaker dependence on $C$ if any. 
In the appendix we show the advantage (or disadvantage) of the \PEW bound over the \PO bound on
\cref{fig:adv}. From this figure we can see that on \pim, \PEW is to be preferred almost uniformly for small values of $C$ ($C\le 0.5$), while on \rin, the advantage of \PEW is limited both for smaller values of $C$ and a certain range of the RBF width. Two comments are in order in connection to this:
{\em (i)} We find it remarkable that a stability-based bound can be competitive with the \PO bound, which is known as one of the best bounds available. \todoc{citation}
{\em (ii)} While comparing bounds is interesting for learning about their qualities, the bounds can be used together (e.g., at the price of an extra union bound).

{\em Model selection}
To evaluate a bounds capability in helping model selection it is worth comparing the correlation between the bound and test error of the underlying classifiers. By comparing 
\cref{fig:poo,fig:testerrpoo}  with  \cref{fig:oour,fig:testerroour} it appears that perhaps the behavior of the \PEW bound (at least for small values of $C$) follows more closely the behavior of the corresponding test error surface. This is particularly visible on \rin, where the \PEW bound seems to be able to pick better values both for $C$ and $\sigma$, which lead to a much smaller test error (around $0.12$) than what one can obtain by using the \PO bound. 

\section{Discussion}

We have developed a PAC-Bayes bound for randomized classifiers.\todoc{Discuss the experiments...}
We proceeded by investigating the stability of the hypothesis
learned by a Hilbert space valued algorithm. A special case being SVMs.
We applied our main theorem to SVMs, leading to our \PEW bound,
and we compared it to other stability-based bounds and to a
previously known PAC-Bayes bound. The main finding is that perhaps \PEW 
is the first nontrivial bound that uses stability.



\bibliography{biblio}

\newpage
\appendix

\section{Proof of \cref{lemma-fbd}}
\label{app:proof-lemma}

Let $M_n = f(Z_1,\ldots,Z_n)$ be a function of
the independent $\mathcal{Z}$-valued random variables $Z_1,\ldots,Z_n$,
where the function $f : \mathcal{Z}^n \to \mathcal{H}$
maps into a separable Hilbert space $\mathcal{H}$.
Let's write
$M_n - \E[M_n]$ as the telescopic sum\footnote{The Doob decomposition: 
$D_i$ are martingale differences and their sum $M_n - \E[M_n]$ is a martingale.}
\begin{align*}
M_n - \E[M_n] = D_n  + D_{n-1} + \cdots + D_{1}
\end{align*}
where
\begin{align*}
D_i = \E[M_n | \mathcal{F}_i] - \E[M_n | \mathcal{F}_{i-1}]
\end{align*}
and $\mathcal{F}_k = \sigma(Z_1,\ldots,Z_k)$
the $\sigma$-algebra generated by the first $k$ examples.
Thus
\begin{align*}
\left\Vert M_n - \E[M_n] \right\Vert^2
= \sum_{i=1}^{n} \norm{D_i}^2 + 2\sum_{i<j} \ip{D_i}{D_j}\,.
\end{align*}
We need $\E\norm{M_n - \E[M_n]}^2$.
Taking the expectation above makes the second sum disappear
since for $i<j$ we have
\begin{align*}
\E[\ip{D_i}{D_j}]
= \E\bigl[ \E[\ip{D_i}{D_j} | \mathcal{F}_i] \bigr]
= \E\bigl[ \ip{D_i}{\E[D_j  | \mathcal{F}_i]} \bigr]
\end{align*}
and clearly $\E[D_j  | \mathcal{F}_i] = 0$ for $j > i$.
Thus we have
\begin{equation}
\EE \left\Vert M_n - \E[M_n] \right\Vert^2
= \EE \sum_{i=1}^{n} \norm{D_i}^2\,.
\label{eq-something}
\end{equation}

Also recall the notation
$\xi_{k:l} = (\xi_k,\ldots,\xi_{l})$ for $k < l$.
It will be used extensively in what follows.

Let's write the conditional expectations in terms of
regular conditional probabilities:
\begin{align*}
\E[f(Z_{1:n}) | \mathcal{F}_i]
= \int f(Z_{1:i},z_{i+1:n}) \dee P_{Z_{i+1:n}|Z_{1:i}}(z_{i+1:n}|Z_{1:i}) \,.
\end{align*}
The random variables are labelled with capitals.
The lower case letters are for the variables of integration.
We write $P_{X}$ for the distribution (probability law) of $X$.

Similarly
\begin{align*}
\E[f(Z_{1:n}) | \mathcal{F}_{i-1}] 
&= \int f(Z_{1:i-1},z_{i:n}) \dee P_{Z_{i:n}|Z_{1:i-1}}(z_{i:n}|Z_{1:i-1}) \\
&= \int f(Z_{1:i-1},z_{i:n}) \dee P_{Z_{i}|Z_{1:i-1}}(z_{i}|Z_{1:i-1})
		\cdot \dee P_{Z_{i+1:n}|Z_{1:i}}(z_{i+1:n}|Z_{1:i-1},x_i)\,.
\end{align*}
By independence, $P_{Z_{i+1:n}|Z_{1:i}} = P_{Z_{i+1:n}}$
and $P_{Z_{i}|Z_{1:i-1}} = P_{Z_{i}}$ 
(this latter is not really needed in the proof, but shortens the formulae).
Hence,
\begin{align*}
D_i = \E[f(Z_{1:n}) | \mathcal{F}_i] - \E[f(Z_{1:n}) | \mathcal{F}_{i-1}]
&= \int f(Z_{1:i},z_{i+1:n}) \dee P_{Z_{i+1:n}}(z_{i+1:n})\\
&\hspace{10mm} - \int f(Z_{1:i-1},z_{i:n}) \dee P_{Z_{i}}(z_{i}) 
			\dee P_{Z_{i+1:n}}(z_{i+1:n})\,.
\end{align*}
Then $D_i$ 
is equal to the integral w.r.t. $P_{Z_{i+1:n}}(z_{i+1:n})$
of
\begin{align*}
\int [f(Z_{1:i-1},Z_i,z_{i+1:n}) - f(Z_{1:i-1},z_i,z_{i+1:n})] 
\dee P_{Z_{i}}(z_{i})\,.
\end{align*}
Notice that in the integrand, only the $i$th argument differs.
If $\norm{f(Z_{1:i-1},Z_i,z_{i+1:n}) - f(Z_{1:i-1},z_i,z_{i+1:n})} \leq c$,
then $\norm{D_i} \leq c$. Thus bounded differences for $f(Z_{1:n})$
implies bounded martingale differences (in norm).

Finally, using Jensen's inequality and \eqref{eq-something},
and the bounded differences assumption:
\begin{align*}
\E\norm{ M_n - \E[M_n] }
\leq \sqrt{\E\norm{ M_n - \E[M_n] }^2}
\leq \sqrt{\sum_{i=1}^{n} c_i^2}\,.
\end{align*}

\section{The average empirical error for Gaussian random linear classifiers}

Let $Q = \mathcal{N}(w_0,I)$, 
a Gaussian with center $w_0 \in \mathbb{R}^d$
and covariance matrix the identity $d \times d$.
The average empirical error is to be calculated as
\begin{equation}
\label{eq-average-error}
R(Q,P_n) 
= \int_{\mathcal{X}\times\mathcal{Y}}%
\tilde{F} \left( \frac{y \ w_0^{\top} \phi(x) }{\|\phi(x)\|} \right) \dee P_n(x,y)
\end{equation} 
where $\tilde{F}  = 1- F$ and $F$ is the standard normal cumulative distribution
\begin{equation}
\label{eq-GaussianCDF}
F(x) = \int_{-\infty}^x {\frac{1}{\sqrt{2\pi}}e^{-u^2/2} \dee u}\,.
\end{equation}
Recall that $P_n = \frac{1}{n}\sum_{i=1}^{n} \delta_{(X_i,Y_i)}$ is
the empirical measure on $\mathcal{X}\times\mathcal{Y}$ 
associated to the $n$ training examples,
and the integral with respect to $P_n$ evaluates as a normalized sum.

In this section we write the derivation of \eqref{eq-average-error}.

To make things more general let $Q = \mathcal{N}(w_0,\Sigma)$, 
a Gaussian with center $w_0 \in \mathbb{R}^d$
and covariance matrix $\Sigma$.
We'll write $G_{(w_0,\Sigma)}$ for the corresponding 
Gaussian measure on $\mathbb{R}^d$.
But to make notation simpler, lets work with input vectors $x$
(instead of feature vectors $\phi(x) \in \mathcal{H}$).
This is in the context of binary classification, 
so the labels are $y \in \{ \pm 1 \}$.
The classifier $c_{w}(\cdot) = \sign(\langle w,\cdot \rangle)$ 
is identified with the weight vector $w$.
The loss on example $(x,y)$ can be written as
\begin{align*}
\ell(c_{w}(x),y) = \mathbf{1}(c_{w}(x) \neq y) = \frac{1-\sign(y \langle w,x \rangle)}{2}\,.
\end{align*}
We'll talk about the empirical error of $w$, namely
$R(w,P_n) = \int_{\mathcal{X}\times\mathcal{Y}} \mathbf{1}(c(x) \neq y) \dee P_n(x,y)$.
The average empirical error when choosing a random weight $W$ according to $Q$ is:
\begin{align*}
R(Q,P_n)
= \int_{\R^d} R(w,P_n) \dee G_{(w_0,\Sigma)}(w)\,.
\end{align*}
Plugging in the definition of $R(w,P_n)$ and swapping the order of the integrals
and using the above formula for the loss, the right hand side is
\begin{align*}
\int_{\R^d} \int_{\mathcal{X}\times\mathcal{Y}} \mathbf{1}(c(x) \neq y) \dee P_n(x,y) & \dee G_{(w_0,\Sigma)}(w) \\
&= \int_{\mathcal{X}\times\mathcal{Y}}\int_{\R^d}  \mathbf{1}(c(x) \neq y) \dee G_{(w_0,\Sigma)}(w) \dee P_n(x,y) \\
&= \int_{\mathcal{X}\times\mathcal{Y}}\int_{\R^d}  \frac{1-\sign(y \langle w,x \rangle)}{2} \dee G_{(w_0,\Sigma)}(w) \dee P_n(x,y) \\
&= \int_{\mathcal{X}\times\mathcal{Y}} \frac{1}{2}\left( 1 - A(x,y) \right) \dee P_n(x,y)
\end{align*}
where for a fixed pair $(x,y)$ we are writing
\begin{align*}
A(x,y) = \int_{\R^d}\sign(y \langle w,x \rangle) \dee G_{(w_0,\Sigma)}(w)\,.
\end{align*}
Decompose into two terms:
\begin{align*}
A(x,y)
= \int_{y\langle w,x \rangle > 0} \!\!\! \dee G_{(w_0,\Sigma)}(w) 
		- \int_{y\langle w,x \rangle < 0} \!\!\! \dee G_{(w_0,\Sigma)}(w)
\end{align*}
and notice that for the random vector $W \sim \mathcal{N}(w_0,\Sigma)$
we have $\E[y\langle W,x \rangle] = y\langle w_0,x \rangle$ and
$\E[(y\langle W,x \rangle)^2] = \Vert x \Vert_{\Sigma}^2 + (\langle w_0,x \rangle)^2$,
so the functional $y\langle W,x \rangle$ has a 1-dimensional Gaussian distribution
with mean $y\langle w_0,x \rangle$ 
and variance $\Vert x \Vert_{\Sigma}^2 = \langle \Sigma x,x \rangle$.
Then
\begin{align*}
\int_{y\langle w,x \rangle > 0} \dee G_{(w_0,\Sigma)}(w) 
&= \operatorname{Pr}[y\langle W,x \rangle > 0] \\
&= \operatorname{Pr}\left[ 
	\frac{y\langle W,x \rangle - y\langle w_0,x \rangle}{\Vert x \Vert_{\Sigma}}%
     > \frac{- y\langle w_0,x \rangle}{\Vert x \Vert_{\Sigma}} \right] \\
&= \operatorname{Pr}\left[ \mathcal{N}(0,1)%
     > \frac{- y\langle w_0,x \rangle}{\Vert x \Vert_{\Sigma}} \right] \\
&= \operatorname{Pr}\left[ \mathcal{N}(0,1)%
     < \frac{y\langle w_0,x \rangle}{\Vert x \Vert_{\Sigma}} \right]
= F\left( \frac{y\langle w_0,x \rangle}{\Vert x \Vert_{\Sigma}} \right)\,.
\end{align*}
Then
\begin{align*}
A(x,y)
= 2F\left( \frac{y\langle w_0,x \rangle}{\Vert x \Vert_{\Sigma}} \right) - 1
\end{align*}
and
\begin{align*}
1-A(x,y)
= 2 - 2F\left( \frac{y\langle w_0,x \rangle}{\Vert x \Vert_{\Sigma}} \right)
= 2\tilde{F}\left( \frac{y\langle w_0,x \rangle}{\Vert x \Vert_{\Sigma}} \right)\,.
\end{align*}
Altogether this gives
\begin{align*}
R(Q,P_n)
= \int_{\mathcal{X}\times\mathcal{Y}}%
\tilde{F}\left( \frac{y\langle w_0,x \rangle}{\Vert x \Vert_{\Sigma}} \right) \dee P_n(x,y)\,.
\end{align*}
Notice that using $\Sigma = I$ (identity) and $\phi(x)$ instead of $x$ 
this gives \eqref{eq-average-error}.

REMARK: \citet{Langford05} uses a $Q$ which is $\mathcal{N}(\mu,1)$ 
along the direction of a vector $w$, 
and $\mathcal{N}(0,1)$ in all directions perpendicular to $w$.
Such $Q$ is a Gaussian centered at $w_{0} = \mu w/\Vert w \Vert$,
giving his formula
\begin{equation*}
R(Q,P_n) 
= \int_{\mathcal{X}\times\mathcal{Y}}%
\tilde{F} \left( \mu \frac{y \ w^{\top} \phi(x) }{\|w\| \ \|\phi(x)\|} \right) 
\dee P_n(x,y)\,.
\end{equation*}

\section{SVM weight vector: clarification about formulations}
\label{app:SVMformulations}

We have a sample of size $n$.

In the standard implementation the weight vector $W_{n}(C)$
found by $\operatorname{SVM}$
is a solution of the following optimization problem:
\begin{equation*}
W_{n}(C) = 
\argmin_{w} \biggl( \frac{1}{2} \norm{w}_{\mathcal{H}}^2 + C \sum_{i=1}^{n} \xi_i \biggr)\,.
\tag{svm1}
\label{eq:svm1}
\end{equation*}
In our paper the weight vector $W_{n}^{\textrm{OURS}}(\lambda)$
found by $\operatorname{SVM}$
is a solution of the following optimization problem:
\begin{equation*}
W_{n}^{\textrm{OURS}}(\lambda) = 
\argmin_{w} \biggl( \frac{\lambda}{2} \norm{w}_{\mathcal{H}}^2 + \frac{1}{n}\sum_{i=1}^{n} \xi_i \biggr)\,.
\tag{svm2}
\label{eq:svm2}
\end{equation*}
In \citet{BE2002Stability} and \citet{Liu-etal2017BarcelonaPaper}
the weight vector $W_{n}^{\textrm{B\&E}}(\lambda)$
found by $\operatorname{SVM}$ 
is a solution of the following optimization problem:
\begin{equation*}
W_{n}^{\textrm{B\&E}}(\lambda) = 
\argmin_{w} \biggl( \lambda \norm{w}_{\mathcal{H}}^2 + \frac{1}{n}\sum_{i=1}^{n} \xi_i \biggr)\,.
\tag{svm3}
\label{eq:svm3}
\end{equation*}

The minimum is over $w \in \mathcal{H}$ (an appropriate Hilbert space)
and subject to some constrains for the $\xi_i$'s in all cases.

The relation between them is: 
\begin{itemize}[leftmargin=*, itemsep=5pt, parsep=2pt]
\item $W_{n}^{\textrm{OURS}}(\lambda) = W_{n}^{\textrm{B\&E}}(\lambda/2)$
\item $W_{n}^{\textrm{B\&E}}(\lambda) = W_{n}(C)$ with $C = \frac{1}{2n\lambda}$
\item $W_{n}^{\textrm{OURS}}(\lambda) = W_{n}(C)$ with $C = \frac{1}{n\lambda}$
\end{itemize}

\section{Details for experiments}
\label{sec:expdetails}
In this section we show further details and results that did not fit the main body of the paper.

\subsection{Details of optimizing $\sigma_{\text{noise}}^2$}
This optimization is ``free'' for the \PO bound as the bound is uniform over $\sigma_{\text{noise}}^2$.
In the \PEW bound we adjusted the failure probability $\delta$ to accommodate the multiple evaluations during the optimization  by replacing it with $\delta/(\tau(\tau+1))$, where $\tau$ is the number of times the \PEW bound is evaluated by the optimization procedure. 
A standard union bound argument shows that the adjustment to $\delta$ makes the resulting bound hold with probability $1-\delta$ regardless the value of $\tau$. 
The \textsc{SLSQP} method implemented in \textsc{scipy} was used as an optimizer, with an extra outer loop that searched for a suitable initialization, as \textsc{SLSQP} is a gradient based method and the \PO bound can be quite ``flat''. The same problem did not appear for the \PEW bound. The attentive reader may be concerned that if $\tau$ gets large values, we, in a way, are optimizing the ``wrong bound''. To check whether this is a possibility, we also computed the ``union bound penalty'' for decreasing $\delta$ by the factor $\tau(\tau+1)$ as the difference between the (invalid) bound where $\delta$ is unchanged and the bound where $\delta$ is decreased and found that the penalty was generally orders of magnitudes smaller than the risk estimate. Nevertheless, this may be a problem when the risk to be estimated is very small, which we think is not very common in practice. 

\subsection{Additional figures}

\begin{figure}[h]
\includegraphics[width=0.5\textwidth]{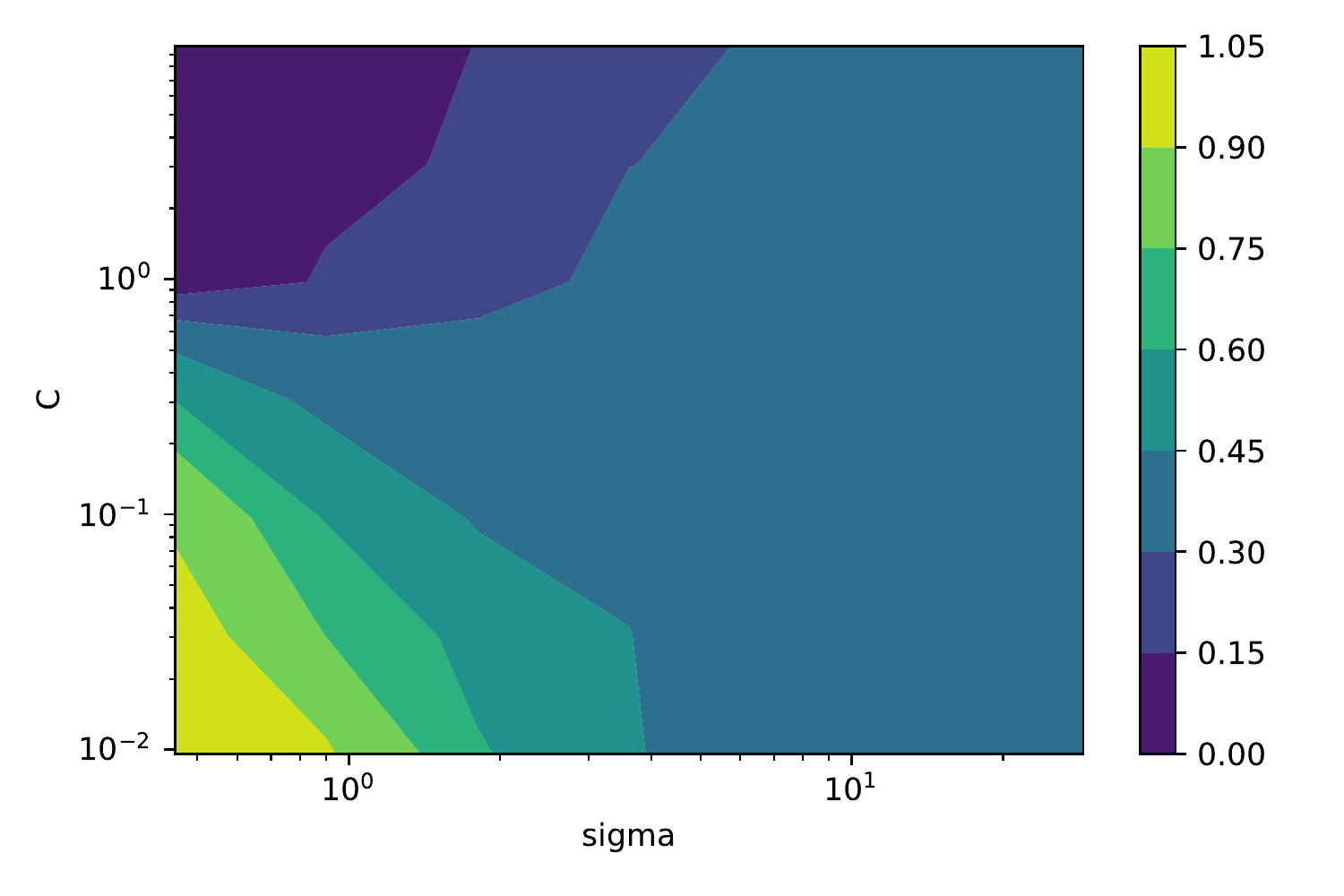}
\includegraphics[width=0.5\textwidth]{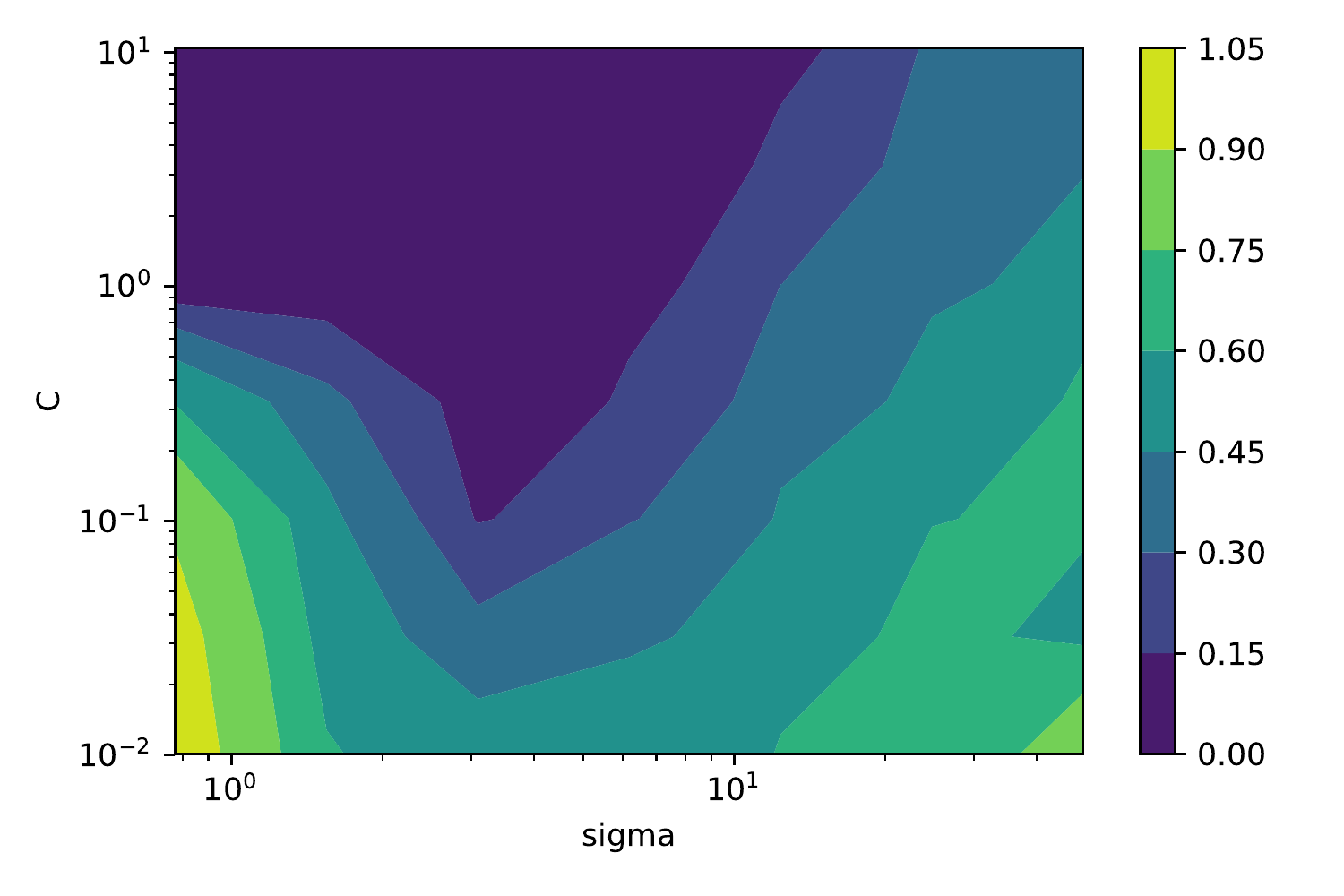}
\caption{Hinge loss on \pim (left) and \rin (right).
For large values of $C$, the hinge loss is reasonable, but this is not the case for small values.
}
\label{fig:hinge}
\end{figure}

\begin{figure}[h]
\includegraphics[width=0.5\textwidth]{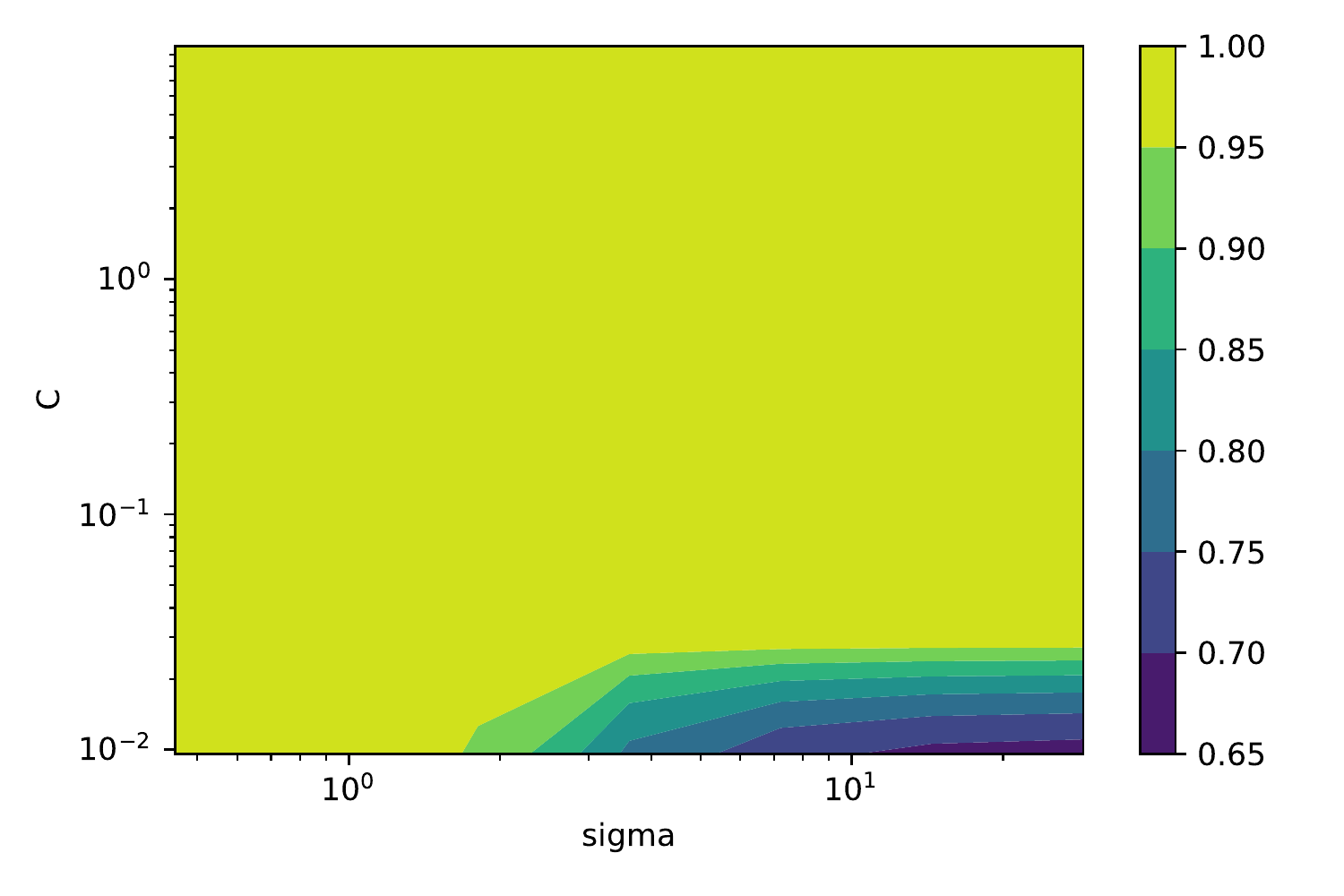}
\includegraphics[width=0.5\textwidth]{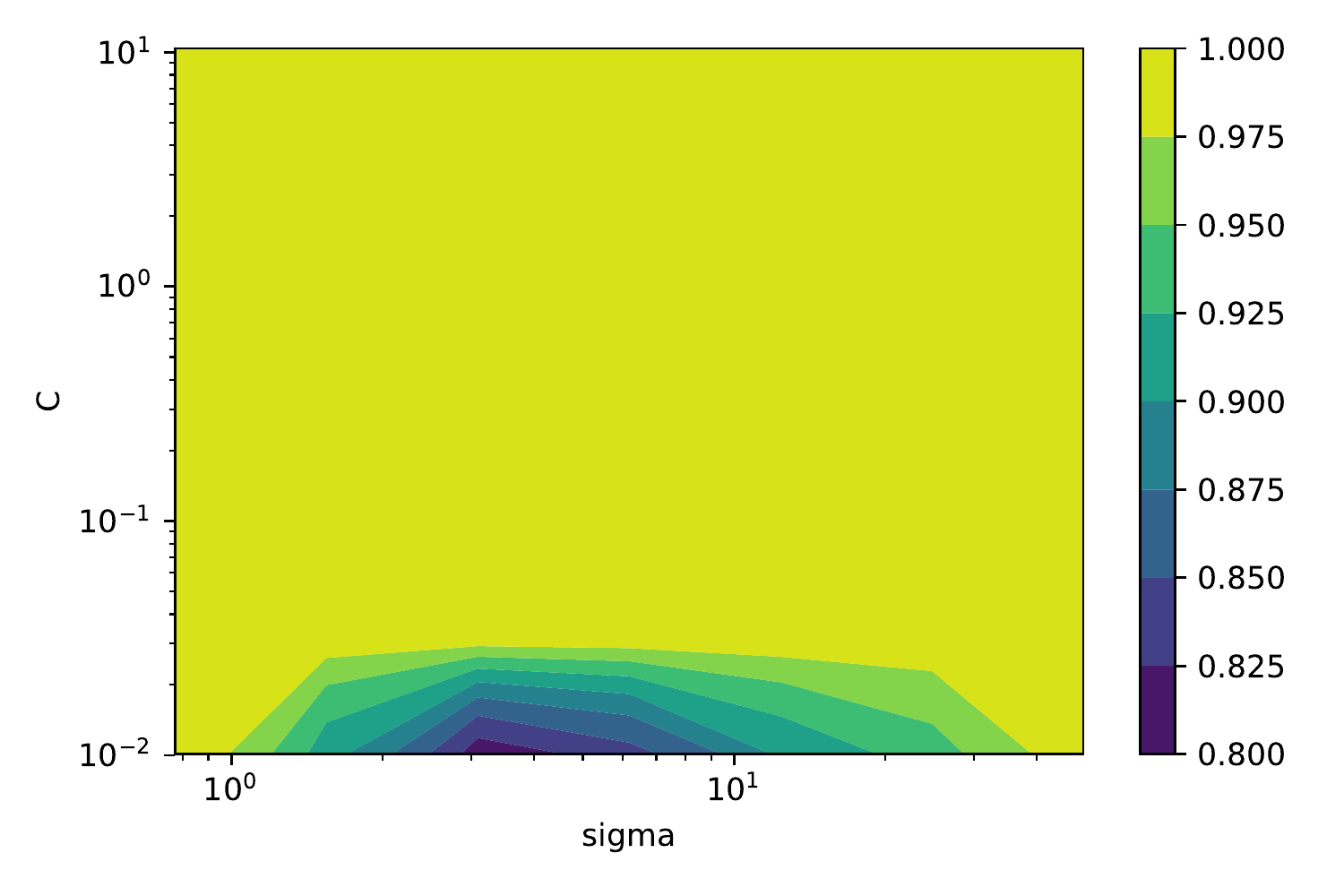}
\caption{The bound of 
\citet{Liu-etal2017SpanishPaper} on \pim (left) and \rin (right).
The bound is almost always vacuous for reasons described in the text.
}
\label{fig:liu}
\end{figure}

\begin{figure}[h]
\includegraphics[width=0.5\textwidth]{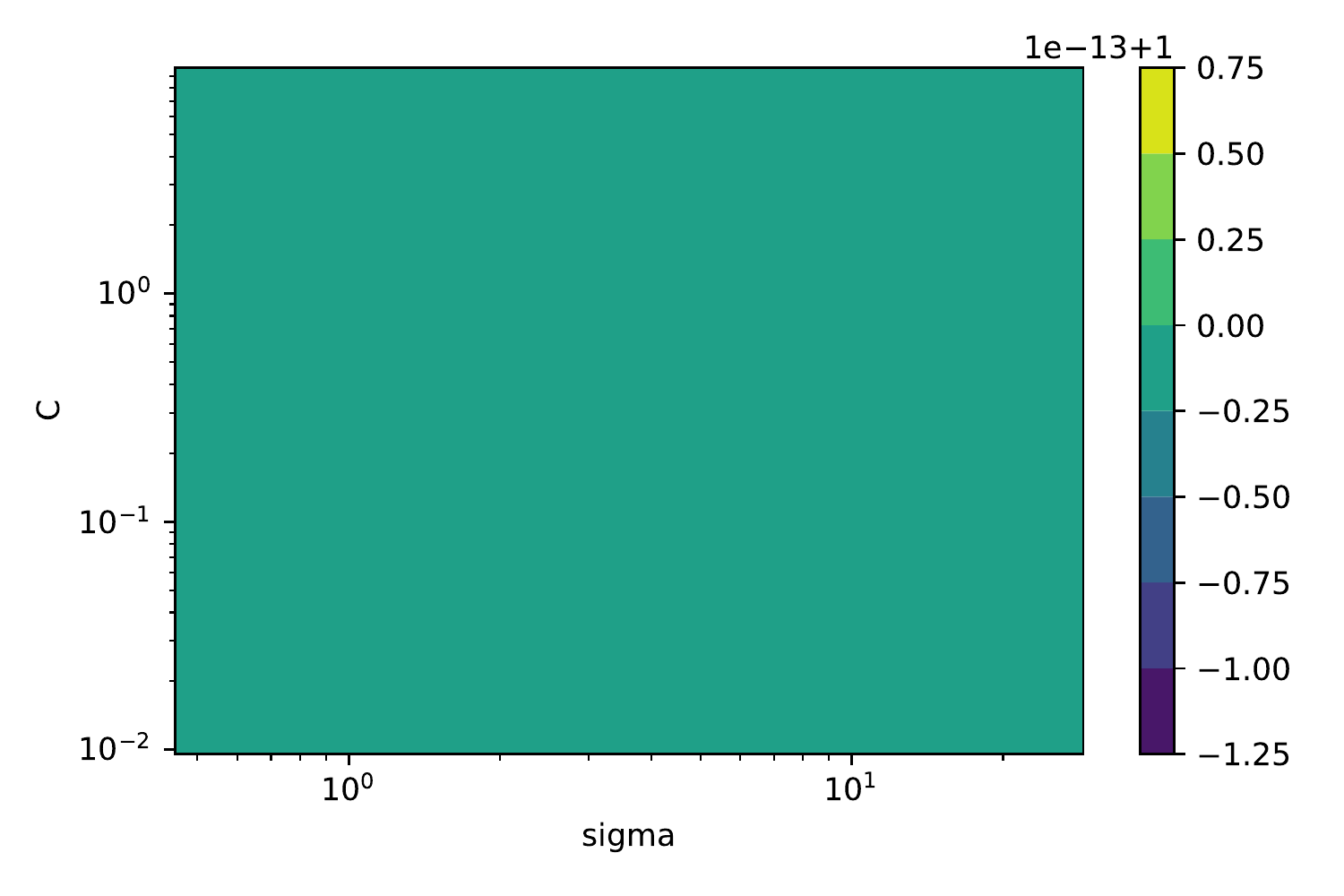}
\includegraphics[width=0.5\textwidth]{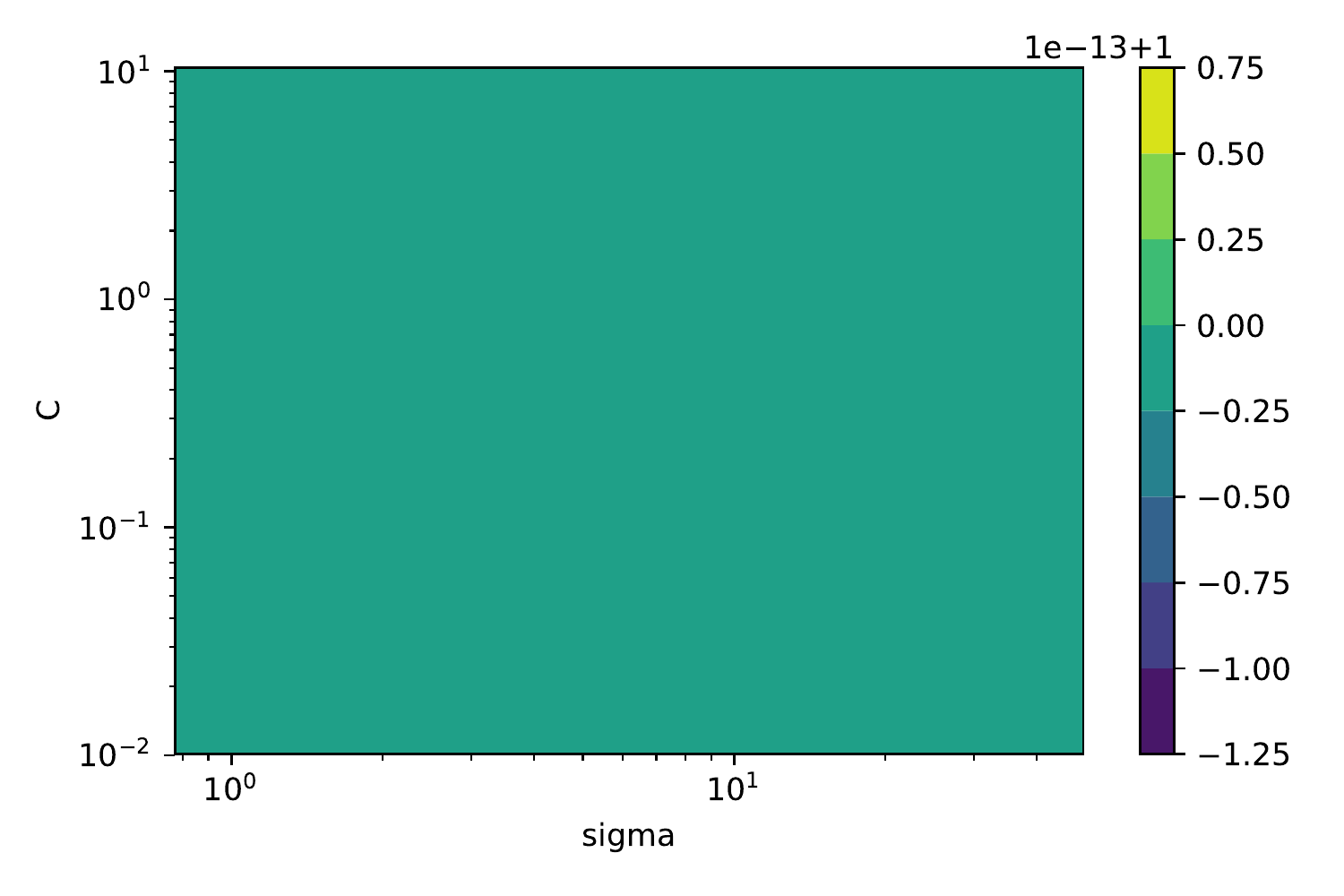}
\caption{The bound of 
\citet{BE2002Stability} on \pim (left) and \rin (right).
The bound is almost always vacuous for reasons described in the text.
}
\label{fig:be}
\end{figure}

\begin{figure}[t]
\includegraphics[width=0.5\textwidth]{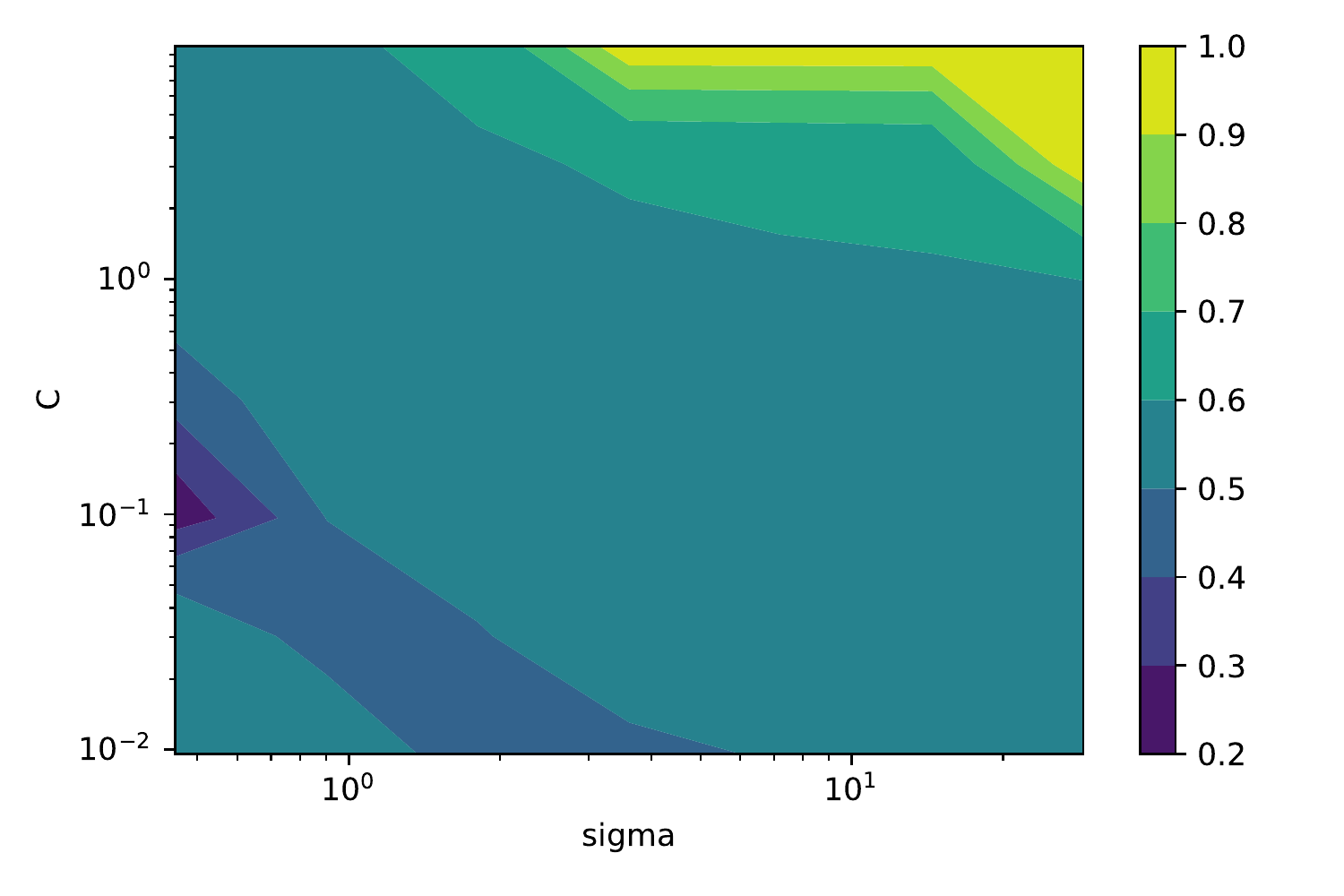}
\includegraphics[width=0.5\textwidth]{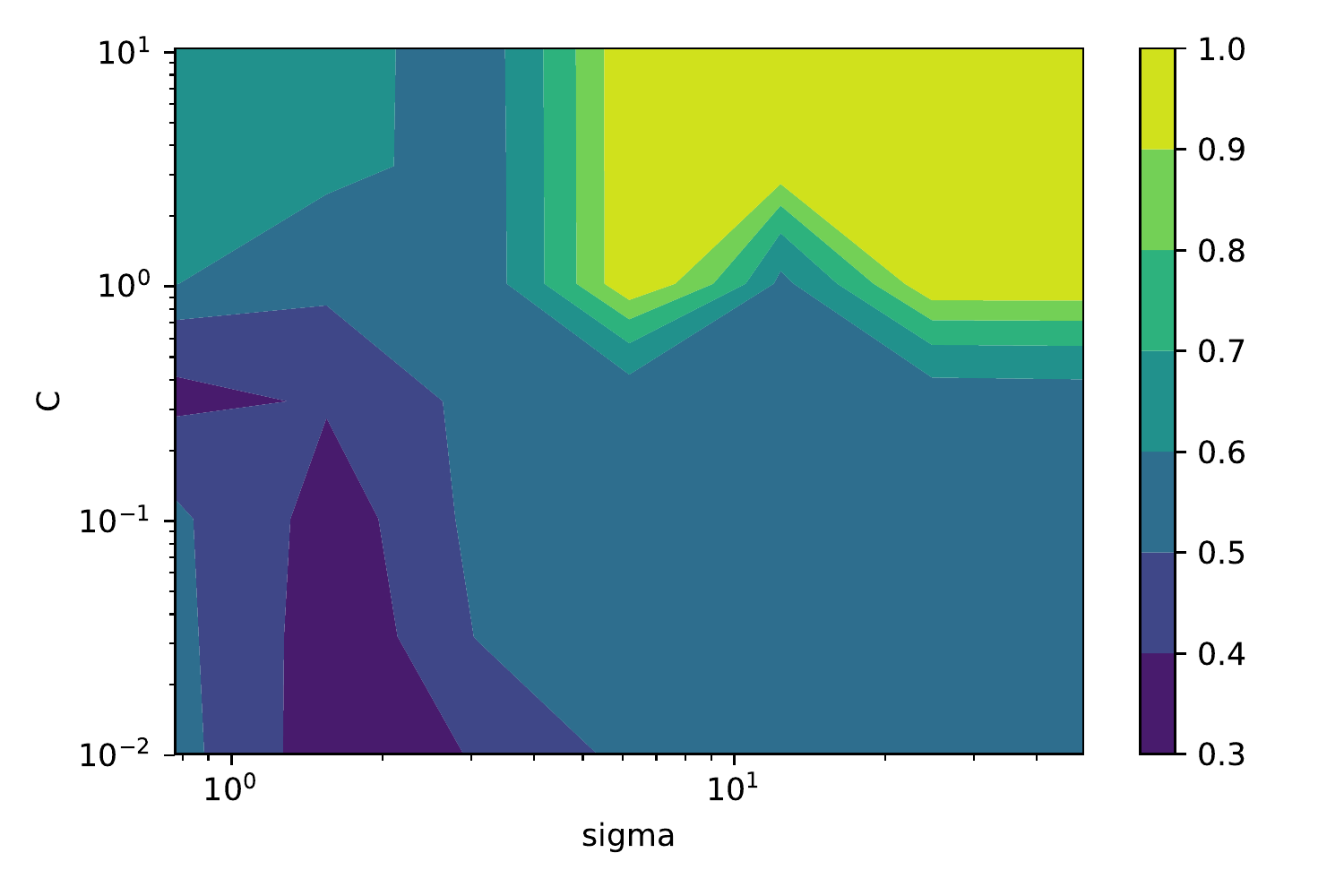}
\caption{The \PO bound on \pim (left) and \rin (right).
}
\label{fig:poo}
\end{figure}

\begin{figure}[t]
\includegraphics[width=0.5\textwidth]{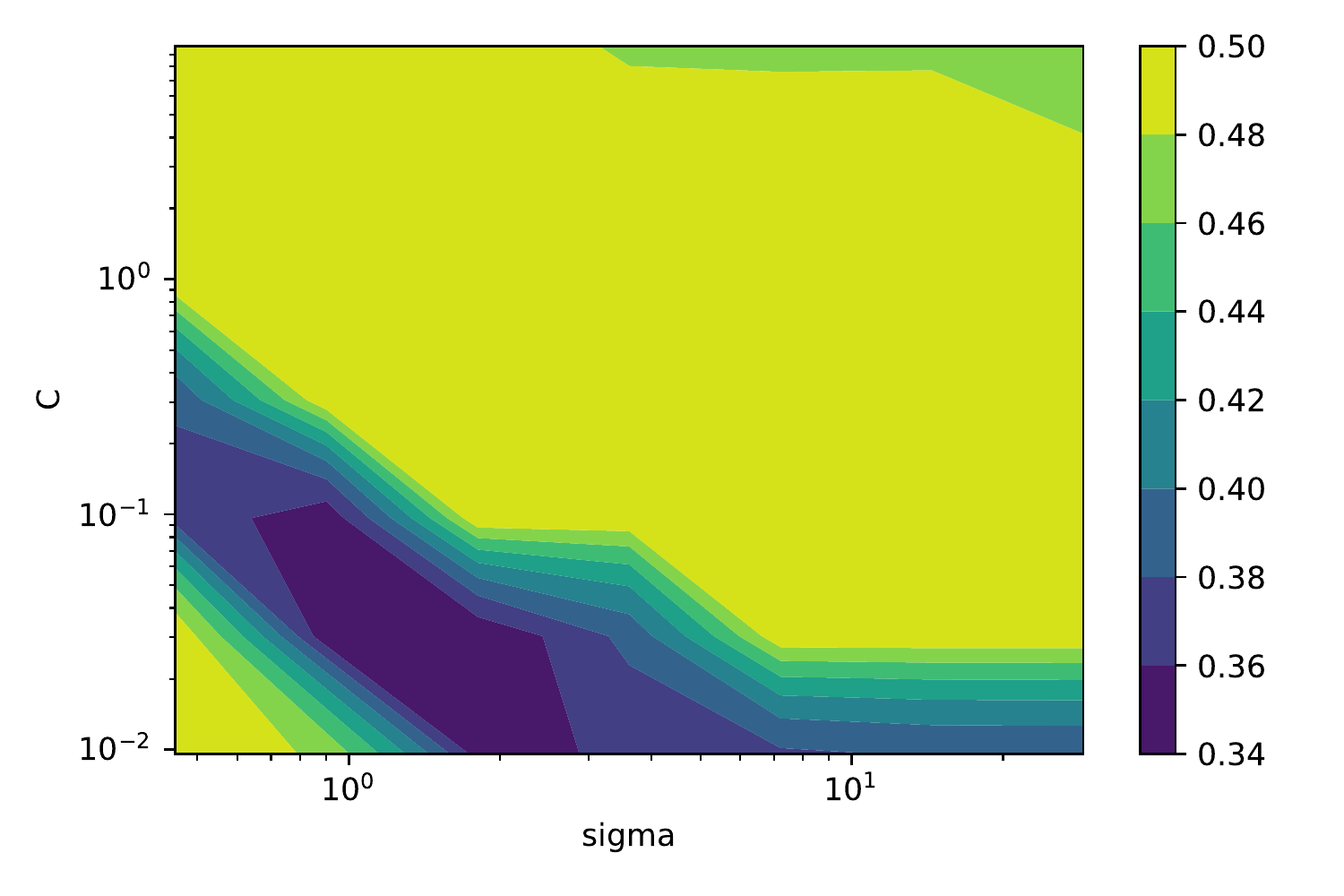}
\includegraphics[width=0.5\textwidth]{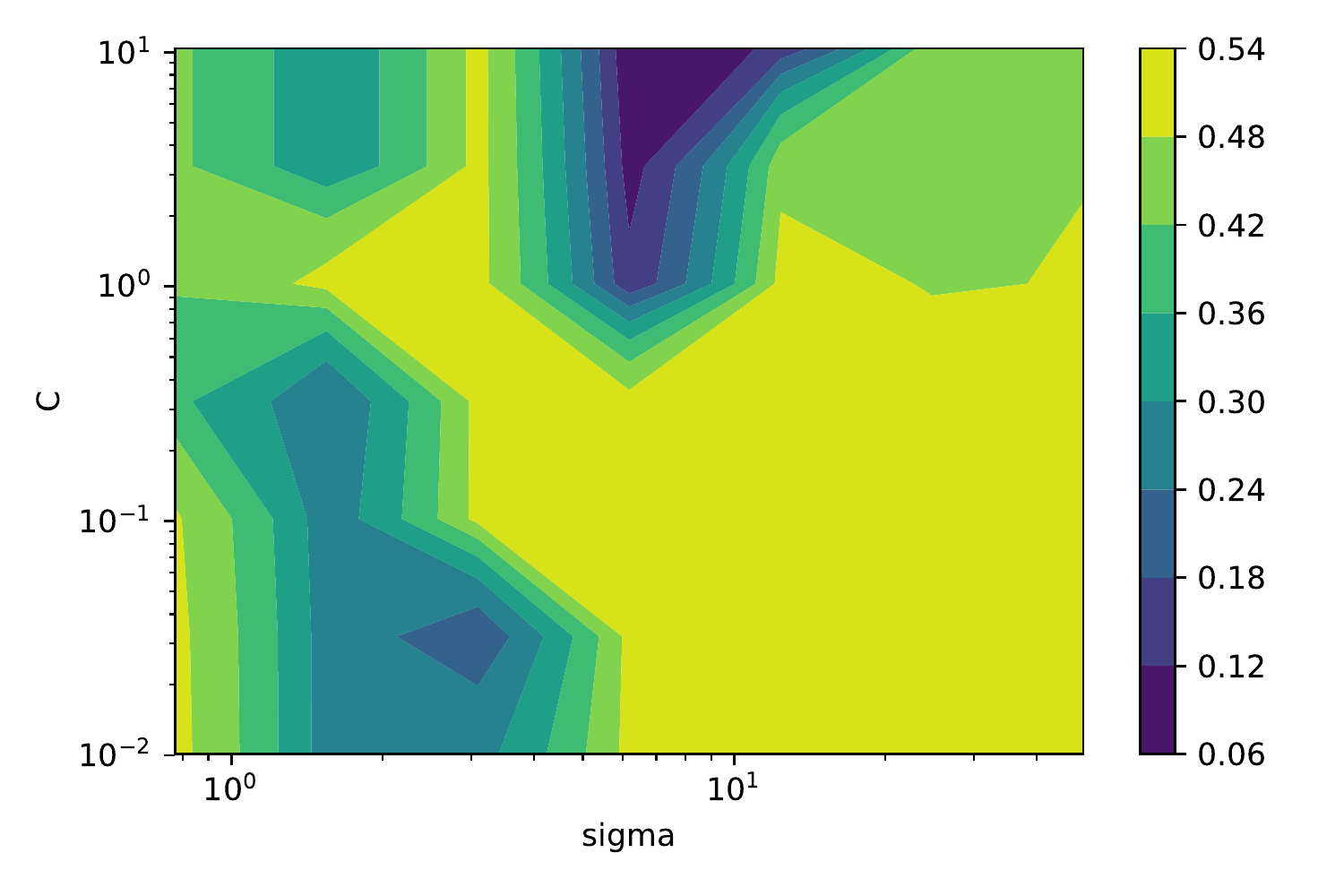}
\caption{Test error of the randomized classifiers underlying the \PO bound on \pim (left) and \rin (right).
}
\label{fig:testerrpoo}
\end{figure}

\begin{figure}[t]
\includegraphics[width=0.5\textwidth]{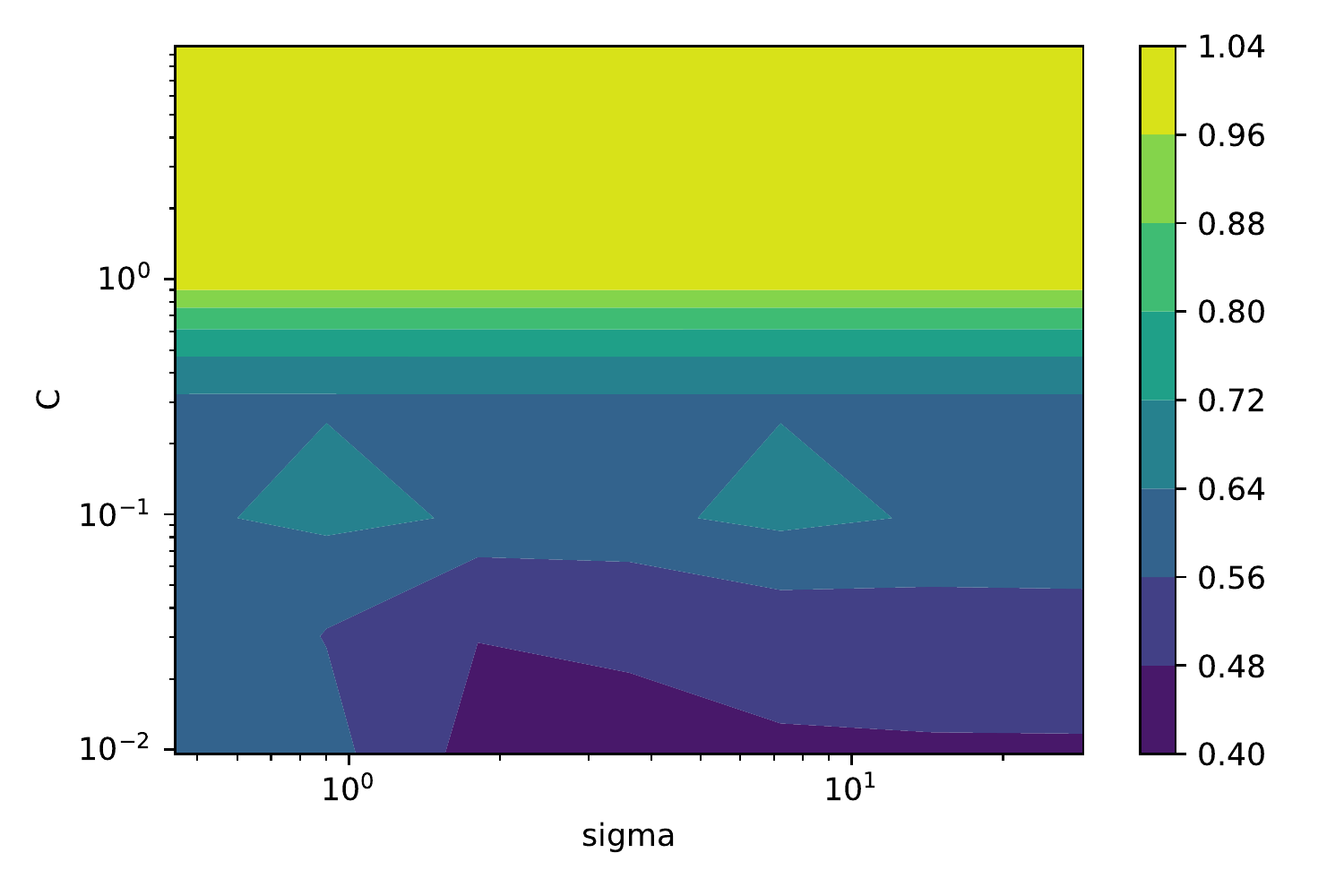}
\includegraphics[width=0.5\textwidth]{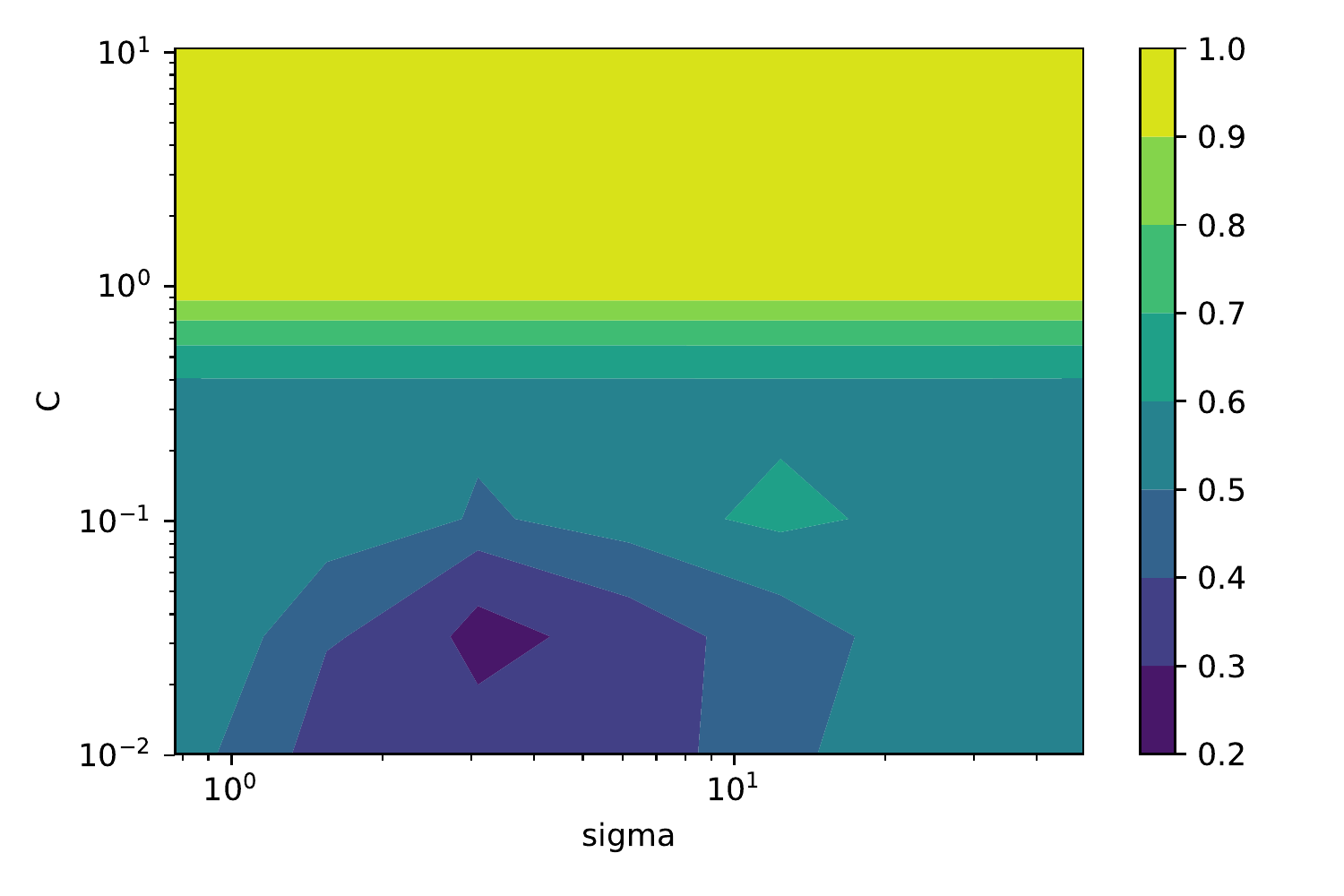}
\caption{The \PEW bound on \pim (left) and \rin (right).
}
\label{fig:oour}
\end{figure}

\begin{figure}[t]
\includegraphics[width=0.5\textwidth]{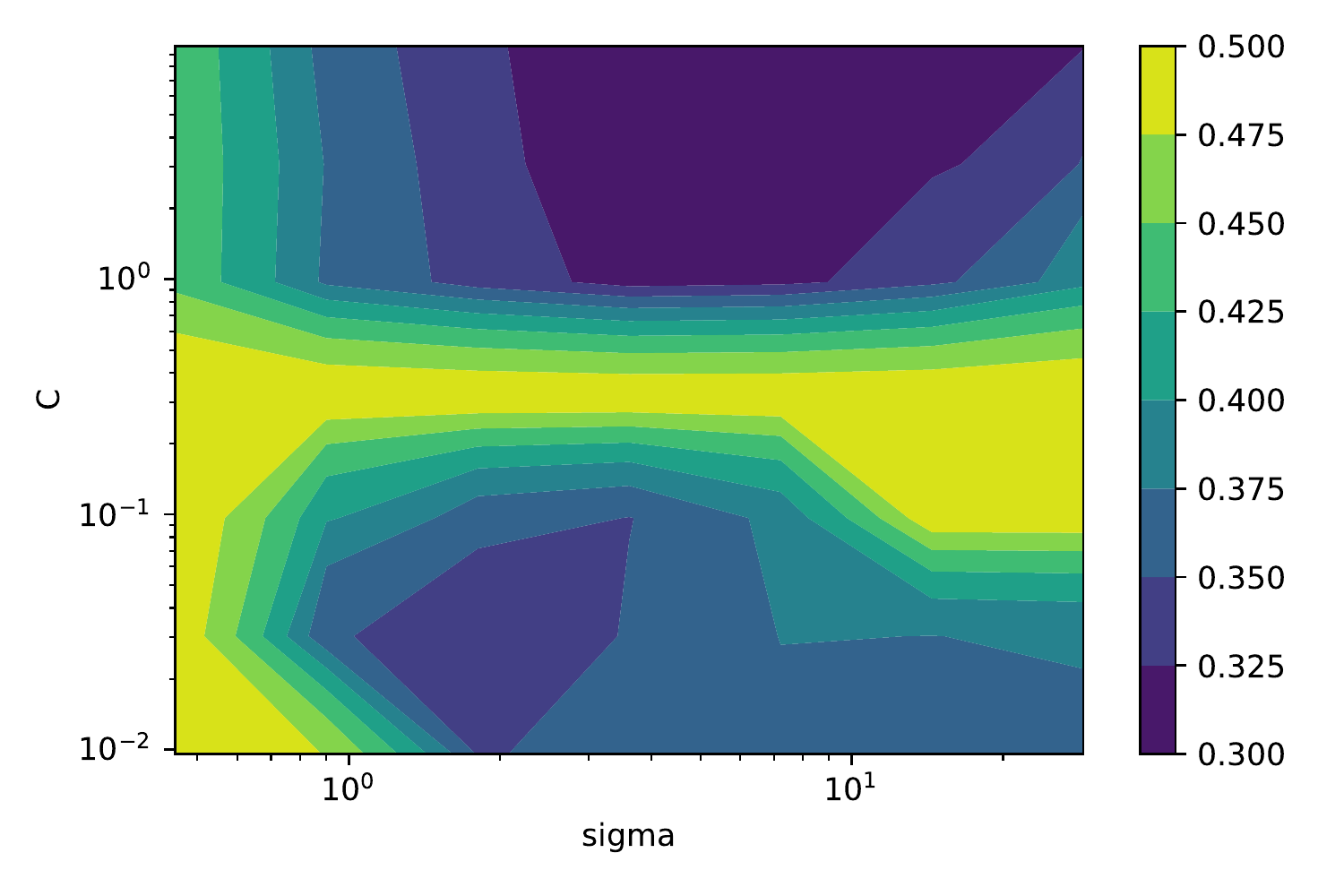}
\includegraphics[width=0.5\textwidth]{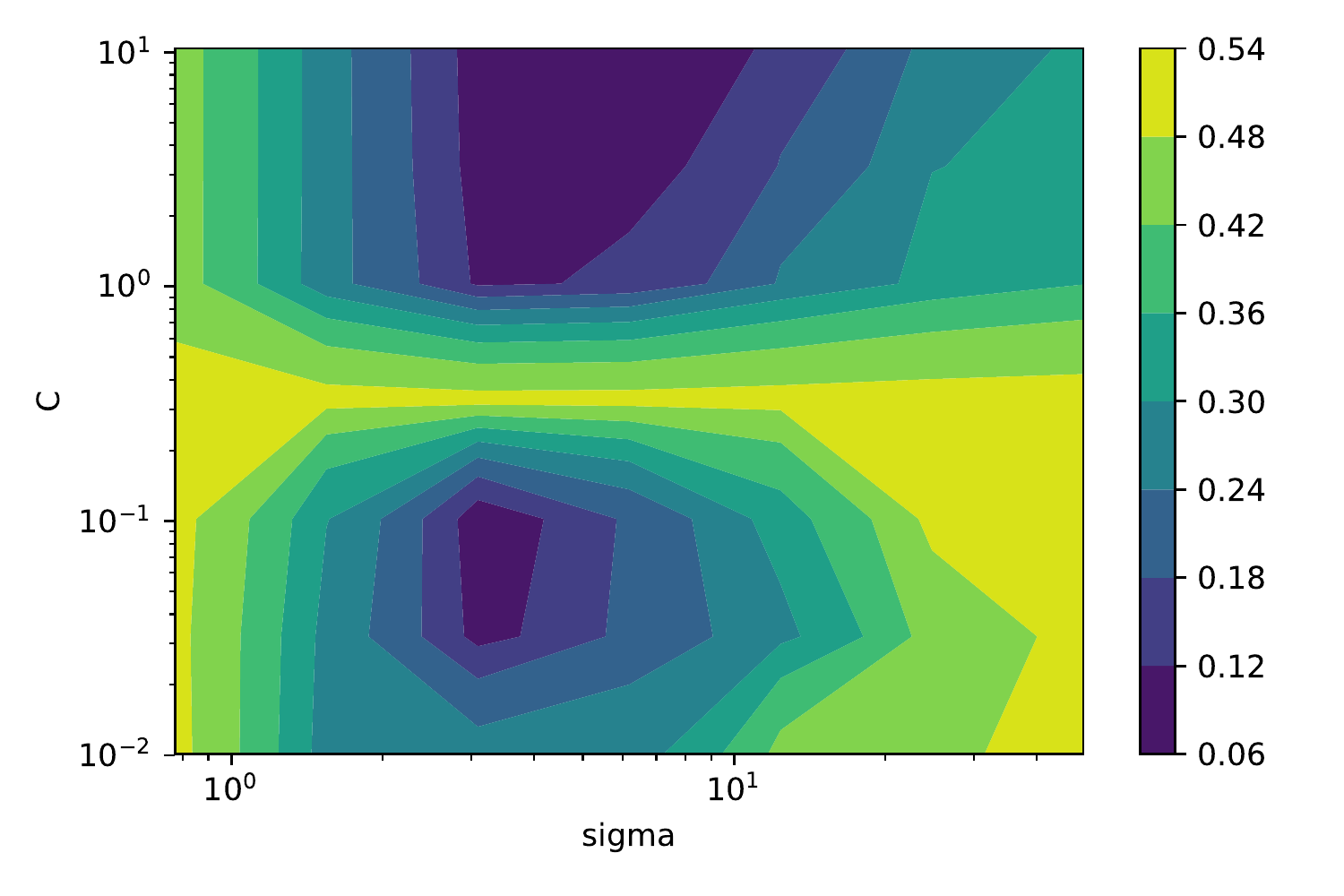}
\caption{Test error of the randomized classifiers underlying the \PEW bound on \pim (left) and \rin (right).
}
\label{fig:testerroour}
\end{figure}

\begin{figure}[t]
\includegraphics[width=0.5\textwidth]{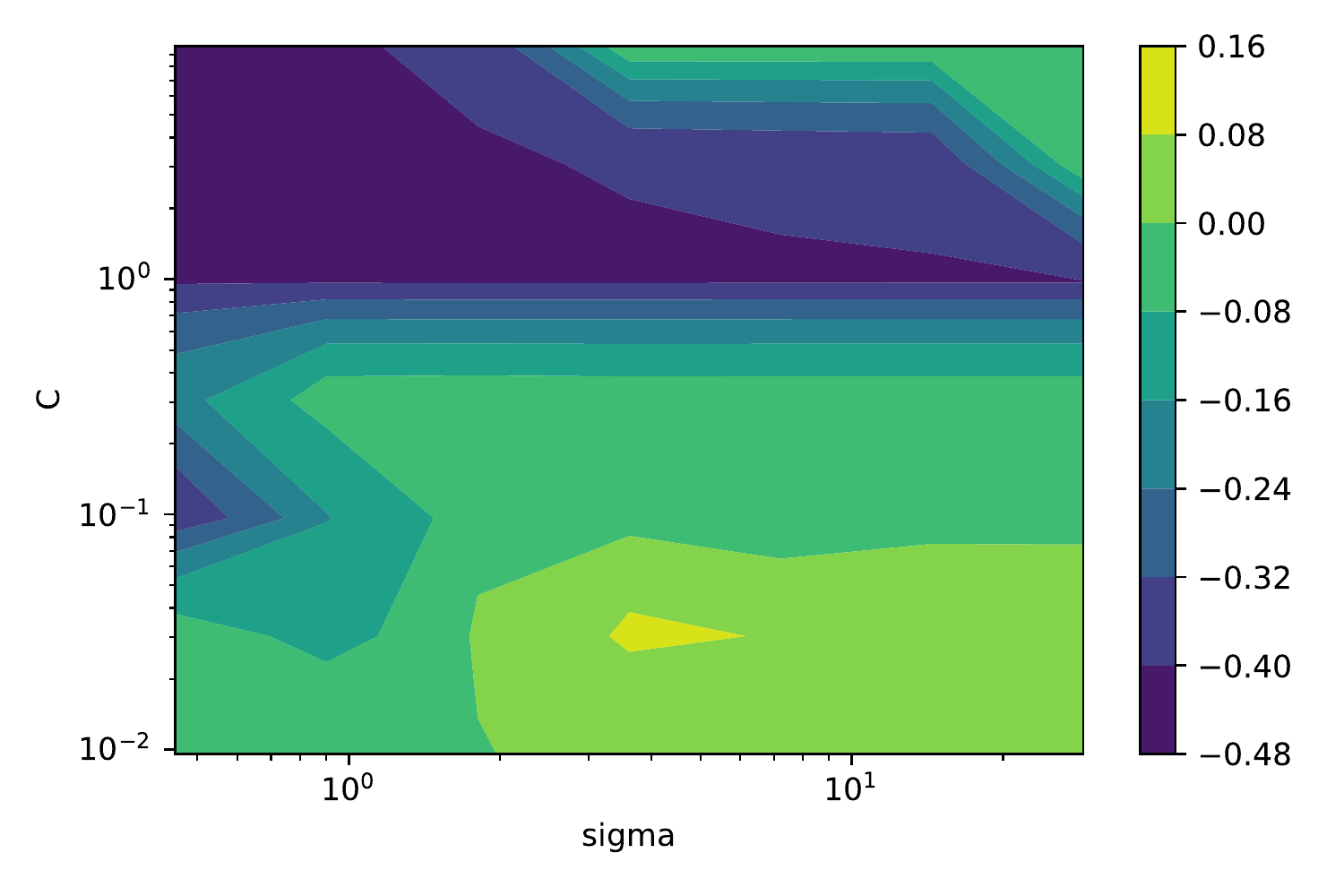}
\includegraphics[width=0.5\textwidth]{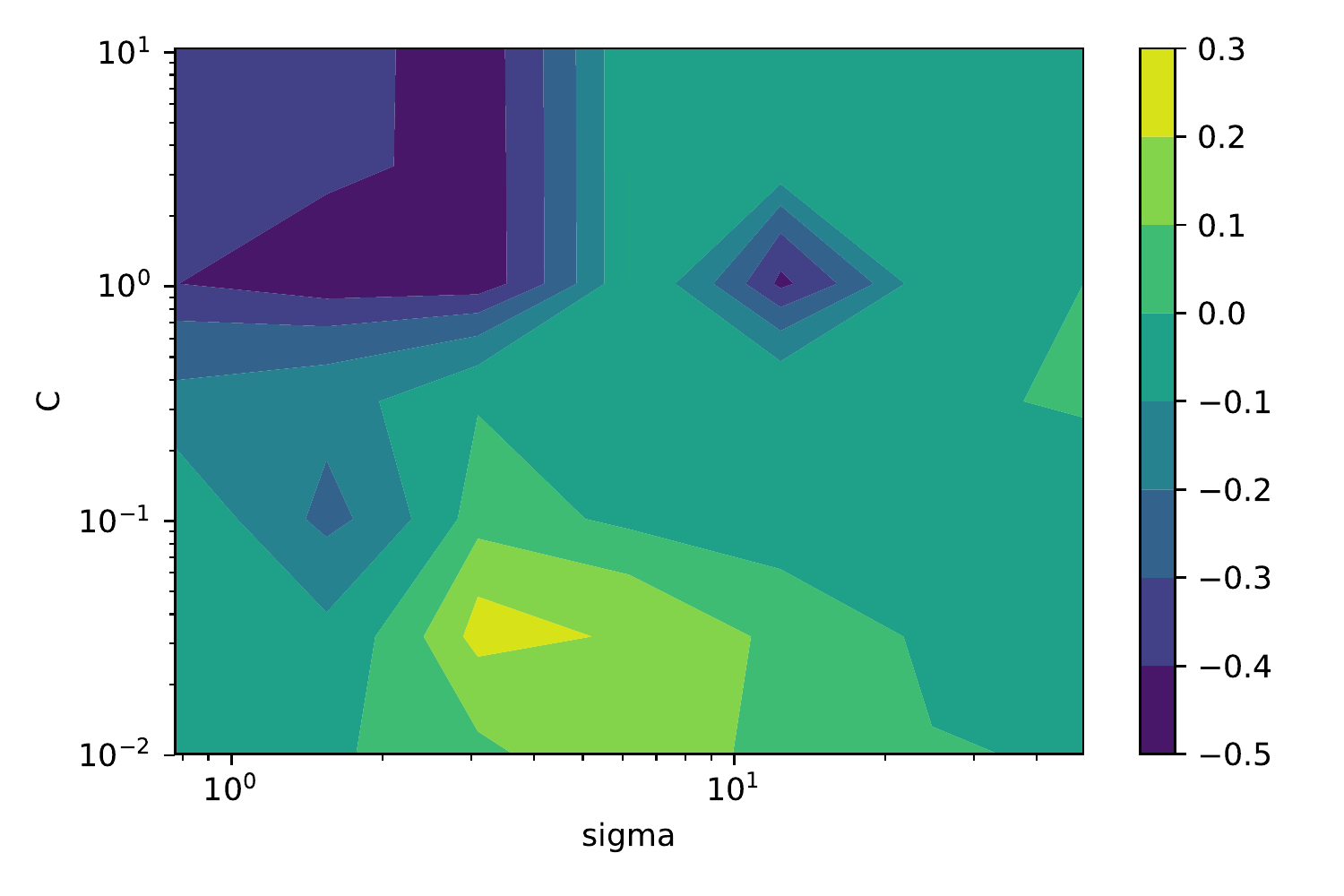}
\caption{Advantage of the \PEW bound to the \PO bound on \pim (left) and \rin (right):
The figure shows the difference between the \PO bound and the \PEW bound.
Where this is positive, \PEW is to be preferred, while where it is negative, \PO is to be preferred.
}
\label{fig:adv}
\end{figure}

\section{Gaussian distributions over the Hilbert space of classifiers?}
\label{app:gaussian}

The idea behind PAC-Bayes is that instead of the weight vector 
$W_n = \operatorname{SVM}(S_n)$ we randomize by choosing a fresh $W \in \mathcal{H}$ 
according to some probability distribution on $\mathcal{H}$ for each prediction.

With the Gaussian kernel in mind, 
we are facing an infinite-dimensional separable $\mathcal{H}$,
which upon the choice of an orthonormal basis $\{ e_1,e_2,\ldots \}$
can be identified with the space $\ell_2 \subset \R^{\N}$
of square summable sequences of real numbers,
via the isometric isomorphism $\mathcal{H} \to \ell_2$ 
that maps the vector $w = \sum_{i=1}^{\infty} w_i e_i \in \mathcal{H}$
to the sequence $(w_1,w_2,\ldots) \in \ell_2$.
Thus without loss of generality we will regard the feature map as
$\phi : \mathcal{X} \to \ell_2 \subset \R^{\N}$.

Suppose the randomized classifier $W$ is to be chosen according to a Gaussian distribution.

Two possibilities come to mind for the choice of random classifier $W$:
(1) according to a Gaussian measure on $\ell_2$,
say $W \sim \mathcal{N}(\mu, \Sigma)$ 
with mean $\mu \in \ell_2$ and covariance operator $\Sigma$ 
meeting the requirements (positive, trace-class) 
for this to be a Gaussian measure on $\ell_2$; or
(2) according to a Gaussian measure on the bigger $\R^{\N}$,
e.g. $W \sim \mathcal{N}(\mu, I)$ by which we mean the measure constructed 
as the product of a sequence $\mathcal{N}(\mu_i,1)$ of independent real-valued Gaussians
with unit variance.
These two possibilities are mutually exclusive
since the first choice gives a measure on $\R^{\N}$ whose mass is supported on $\ell_2$,
while the second choice leads to a measure on $\R^{\N}$ supported outside of $\ell_2$.
A good reference for these things is \cite{Bogachev1998GaussianMeasures}.

Let's go with the second choice:
$\mathcal{N}(0, I) = \bigotimes_{i=}^{\infty} \mathcal{N}(0,1)$,
a `standard Gaussian' on $\R^{\N}$.
This is a legitimate probability measure on $\R^{\N}$
(by Kolmogorov's Extension theorem).
But it
is supported outside of $\ell_2$,
so when sampling a $W \in \R^{\N}$ according to this measure,
with probability one such $W$ will be outside of our feature space $\ell_2$.
Then we have to wonder about the meaning of $\langle W,\cdot \rangle$
when $W$ is not in the Hilbert space carrying this inner product.

Let's write $W = (\xi_1, \xi_2, \ldots)$ a sequence of i.i.d. 
standard (real-valued) Gaussian random variables.
Let $x = (x_1, x_2, \ldots) \in \ell_2$,
and consider the formal expression
$\langle x,W \rangle = \sum_{i=1}^{\infty} x_i \xi_i$. Notice that
\[
\sum_{i=1}^{\infty} \E[ \vert x_i \xi_i \vert^2 ] 
= \sum_{i=1}^{\infty} \vert x_i \vert^2 < \infty\,.
\]
Then (see e.g. \cite{Bogachev1998GaussianMeasures}, Theorem 1.1.4) our formal object
$\langle x,W \rangle = \sum_{i=1}^{\infty} x_i \xi_i$ is actually well-defined 
in the sense that the series is convergent almost surely
(i.e. with probability one), 
although as we pointed out such $W$ is outside $\ell_2$.

\subsection{Predicting with the Gaussian random classifier}

Let $W_n = \operatorname{SVM}(S_n)$ be the weight vector found 
by running SVM on the sample $S_n$. 
We write it as $W_n = \sum_{i=1}^{n} \alpha_i Y_i \phi(X_i)$.

Also as above let $W$ be a Gaussian random vector in $\R^{\N}$,
and write it as
$W = \sum_{j=1}^{\infty} \xi_j e_j$
with $\xi_1,\xi_2,\ldots$ i.i.d. standard Gaussians.
As usual $e_j$ stands for the canonical unit vectors having a 1 in the $j$th coordinate
and zeros elsewhere.

For an input $x \in \mathcal{X}$ 
with corresponding feature vector $\phi(x) \in\mathcal{H}$,
we predict with
\[
\langle W_n + W, \phi(x) \rangle
= \sum_{i=1}^{n} \alpha_i Y_i \kappa(X_i,x) 
	+ \sum_{j=1}^{\infty} \xi_j \langle e_j,\phi(x) \rangle\,.
\]
This is well-defined since
\[
\sum_{i=1}^{\infty} \E[(\xi_j \langle e_j,\phi(x) \rangle)^2] 
= \sum_{i=1}^{\infty} (\langle e_j,\phi(x) \rangle)^2
= \Vert \phi(x) \Vert^2\,,
\]
so the series $\sum_{j=1}^{\infty} \xi_j \langle e_j,\phi(x) \rangle$
converges almost surely (\citet{Bogachev1998GaussianMeasures}, Theorem~1.1.4).

\end{document}